\documentclass{article}

\usepackage[preprint]{neurips_2026}

\usepackage[utf8]{inputenc} %
\usepackage[T1]{fontenc}    %
\usepackage{hyperref}       %
\usepackage{url}            %
\usepackage{booktabs}       %
\usepackage{amsfonts}       %
\usepackage{nicefrac}       %
\usepackage{microtype}      %
\usepackage{xcolor}         %
\usepackage[pdftex]{graphicx}
\usepackage[caption=false]{subfig} %
\usepackage{amsmath}
\usepackage{wrapfig}
\usepackage{algorithm}
\usepackage{algorithmic}

\title{Missing-Modality-Aware Graph Neural Network for Cancer Classification}

\author{
  Sina Tabakhi$^{1}$ \quad
  Chen (Cherise) Chen$^{1}$ \quad
  Haiping Lu$^{1}$ \\
  $^{1}$School of Computer Science, University of Sheffield, Sheffield, UK \\
  \texttt{\{stabakhi1, chen.chen2, h.lu\}@sheffield.ac.uk}
}

\begin{document}

\maketitle

\begin{abstract}
  A key challenge in learning from multimodal biological data is missing modalities, where data from one or more modalities are absent for some patients. Existing approaches either exclude patients with missing modalities, impute missing modalities, or make predictions directly with partial modalities. However, most of these methods rely on inflexible, patient-agnostic fusion strategies and do not scale computationally to the combinatorial growth of missing-modality patterns as the number of modalities increases. To address these limitations, we propose \texttt{MAGNET} (\underline{M}issing-modality-\underline{A}ware \underline{G}raph neural \underline{NET}work) to enhance multimodal prediction with partial modalities, featuring a dynamic \textit{patient-modality multi-head attention} mechanism to fuse lower-dimensional modality embeddings based on their contribution and missingness. \texttt{MAGNET} fusion's complexity increases linearly with the number of modalities while adapting to missing-pattern variability. To generate predictions, \texttt{MAGNET} further constructs a patient graph with fused multimodal embeddings as node features and connectivity determined by the modality missingness, followed by a graph neural network. Experiments on three public multiomics datasets for cancer classification, with \textit{real-world missingness}, show that \texttt{MAGNET} outperforms state-of-the-art fusion methods. The data and code are available at \url{https://github.com/SinaTabakhi/MAGNET}. 
\end{abstract}

\section{Introduction}

Cancer research increasingly profiles patients across multiple molecular layers, known as multiomics, to unravel the complexity of cancer development \cite{swanton2024embracing, su2025interpretable}. By integrating different omics modalities, including DNA methylation, mRNA, and microRNA expression, multimodal machine learning approaches construct comprehensive patient profiles to improve downstream tasks such as cancer classification and subtyping \cite{zitnik2019machine, bi2025msaff, cantini2021benchmarking, acosta2022multimodal}. Conventional multimodal approaches often assume all modalities are available for each patient \cite{wang2014similarity, wang2021mogonet}. However, missing modalities, where all data from one or more modalities are absent for some patients, are unavoidable in biomedical applications \cite{mitra2023learning} due to sample degradation, insufficient data quality, or cost and technical constraints \cite{song2020review, vitrinel2019exploiting}. Therefore, robust multimodal models are needed to handle partial modalities. 

\begin{figure}
  \centering

  \subfloat[Missing modalities follow \textbf{different distributions} across the training and test sets. With $M$ modalities, each patient’s data can have $2^M-1$ possible missing patterns; for the three modalities shown here, seven such patterns are possible. Each colored row shows available patient data, while gray rows indicate missing modalities.]{
    \includegraphics[width=0.48\linewidth]{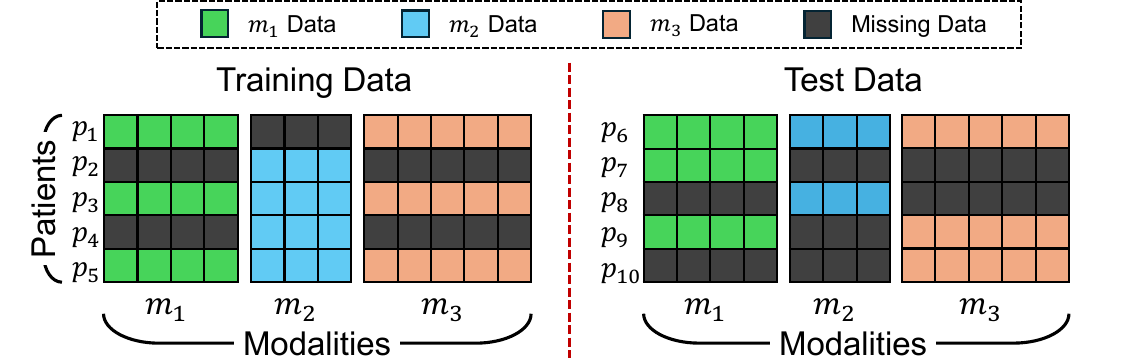}
    \label{fig:miss-pat}
  }
  \hfill
  \subfloat[Existing multimodal approaches either use equal weighting across modalities or apply the same attention weights across modalities or patients. However, in real-world clinical settings, modality importance usually \textbf{differs across patients}. For clarity, missing-modality patterns are not shown in these examples.]{
    \includegraphics[width=0.48\linewidth]{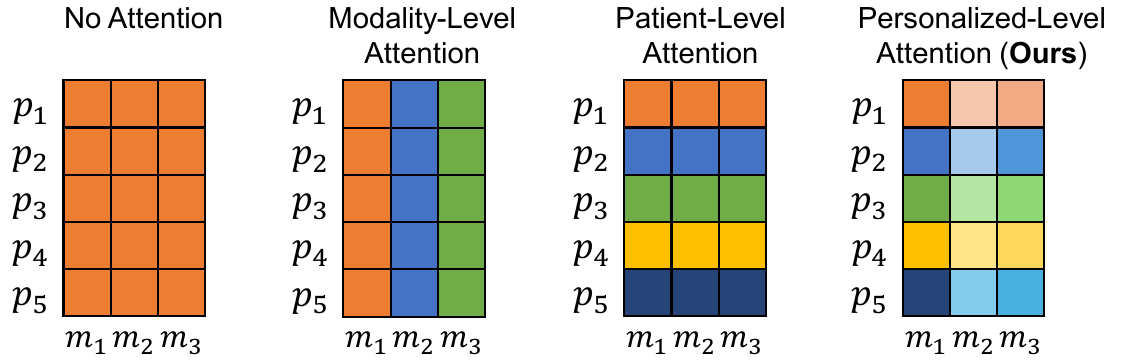}
    \label{fig:att-prob}
  }

  \caption{Challenges of applying direct prediction methods to multimodal biological data.}
  \label{fig:challenges}
\end{figure}

Current approaches to handling missing modalities either exclude patients with missing modalities \cite{wang2020multimodal, wang2021mogonet}, which can significantly reduce the number of available patients \cite{zhang2022m3care, wu2024multimodal}, or impute missingness \cite{collier2020vaes, zhang2022m3care, ngiam2011multimodal}, which may introduce noise \cite{wang2020multimodal} or rely on strong data distribution assumptions \cite{you2020handling}. An alternative approach performs predictions directly with partial modalities \cite{yao2024drfuse, ma2022multimodal}, addressing the limitations of the previous two strategies. We therefore adopt this third approach; however, three challenges remain when applying it to biological data:

\paragraph{Challenge 1: scalability with combinatorial missing-modality patterns} In multiomics cancer studies, missing modalities arise in various patterns across both training and test data due to data collection constraints \cite{wu2024multimodal}. For $M$ modalities, there are $2^M-1$ missing patterns (see Figure~\ref{fig:miss-pat}). Direct methods often construct separate sub-models for each pattern \cite{yun2024flex, yao2024drfuse}, or assume missingness occurs in a single modality \cite{hayat2022medfuse, yao2024drfuse} or only at test time \cite{ma2021smil, reza2024robust}. Such designs do not scale to general missing-modality patterns and therefore present a challenge for cancer profiling.

\paragraph{Challenge 2: inflexible modality fusion} Existing multimodal strategies often assign the same fixed weight to every modality for every patient \cite{yang2023mrgcn}, learn a global attention weight for each modality applied to all patients \cite{ma2022multimodal, caruso2025maria}, or apply attention across patients in a way that treats each patient’s fused representation uniformly \cite{chen2021multimodal, keicher2023multimodal}. However, in clinical settings, modality importance can vary across patients due to clinical heterogeneity \cite{yuan2011patient} (see Figure~\ref{fig:att-prob}).

\paragraph{Challenge 3: artificial missingness evaluation} Evaluating missing-modality handling methods often involves artificially introducing missingness into complete datasets \cite{schouten2018generating}. However, this approach may oversimplify or misrepresent \textit{real-world missingness} patterns \cite{mitra2023learning} and may introduce training-time bias by inadvertently leveraging knowledge of missing modalities during training.

To address these challenges, we propose \texttt{MAGNET} (\underline{M}issing-modality-\underline{A}ware \underline{G}raph neural \underline{NET}work) for direct prediction with partial modalities, with two key contributions:

(1) To tackle challenges 1 and 2, \texttt{MAGNET} introduces a \textit{patient-modality multi-head attention} (PMMHA) mechanism to fuse modalities into a shared multimodal representation by learning patient-specific modality weights. This design allows a \textit{binary modality mask} to handle missing modalities by zeroing out their contributions. While adding a new modality typically introduces an exponential increase in missing-modality patterns, \texttt{MAGNET} requires only one additional learnable attention weight per patient, effectively eliminating the need for separate pattern-specific models. Consequently, \texttt{MAGNET} scales linearly with the number of modalities, keeping the model simple and expandable. To preserve inter-patient similarities during fusion, we introduce a \textit{Kullback–Leibler (KL) divergence-based loss} that minimizes the misalignment between patient similarity distributions before and after fusion, even in the presence of missing modalities. After multimodal fusion, \texttt{MAGNET} further aligns with real-world clinical practice, where doctors often rely on the insight that patients similar in available modalities are likely to share characteristics in missing ones \cite{zhang2022m3care}. Motivated by this principle, we construct a \textit{patient interaction graph} where nodes represent patients with fused representations as their features, and edges connect patients who share at least one available modality, thereby leveraging missing-modality information. A graph neural network (GNN) learns patient representations from this graph to generate final predictions.   

(2) To address challenge 3, we evaluate \texttt{MAGNET} on multiomics datasets for cancer classification with real-world beyond artificial missingness. The results demonstrate that \texttt{MAGNET} consistently outperforms state-of-the-art fusion methods across multiple evaluation metrics.

\section{Related work}

\paragraph{Multimodal learning for cancer} Multiomics profiling has driven the development of multimodal fusion methods in cancer research \cite{kang2022roadmap, reel2021using, zheng2024multi}. Several methods have been proposed to integrate multiomics data \cite{wang2014similarity, chierici2020integrative, schulte2021integration, su2025interpretable, wang2021mogonet}; however, these methods assume that all modalities are available for each patient, limiting their ability to handle missing modalities. Missing modalities are a common challenge in healthcare, caused by low tissue quality, insufficient sample volume, measurement limitations, cost constraints, or patient dropout \cite{flores2023missing, mitra2023learning}.

\paragraph{Multimodal learning with missing modalities} Multimodal models for addressing missing modalities categorize into three approaches. The simplest approach considers only patients with all modalities available, applying methods designed for complete datasets \cite{wang2021mogonet, yang2022deep, wang2014similarity}. While straightforward, this significantly reduces the number of patients and ignores valuable information from those with missing modalities, which is particularly limiting in healthcare settings \cite{pan2021disease}.

Imputation offers an alternative by reconstructing missing modalities using various strategies. One strategy concatenates all modalities, treating missing modalities as missing data and estimating them based on existing values \cite{schafer1999multiple, austin2021missing}. Another strategy uses statistical methods that account for the structure of missing modalities \cite{dong2019tobmi, zhang2022m3care}. Deep generative models such as autoencoders \cite{ngiam2011multimodal, ashuach2023multivi} and generative adversarial networks \cite{cai2018deep} provide another imputation strategy by reconstructing missing modalities from available ones. These models may treat modalities as similar \cite{yao2024drfuse}, introduce imputation noise \cite{zhang2022m3care}, or rely on strong data distribution assumptions \cite{wu2024multimodal}.

Direct prediction is another approach that incorporates specialized designs to perform downstream tasks with missing modalities. NEMO \cite{rappoport2019nemo} manages missing modalities using a neighborhood-based approach with average similarities. MRGCN \cite{yang2023mrgcn} uses a GCN-based encoder-decoder method with an indicator matrix to address missing modalities. GRAPE \cite{you2020handling} and MUSE \cite{wu2024multimodal} leverage bipartite graphs with modalities and patients as nodes to directly address missing modalities. DrFuse \cite{yao2024drfuse} combines modality-specific and shared sub-models. 
However, these methods may oversimplify relationships, struggle with multimodal missingness scalability due to exponentially growing missing patterns, or address only specific missing scenarios \cite{zhang2022m3care, wu2024multimodal}.

\section{Methodology}\label{sec:framework}

\subsection{Problem formulation}

Supervised multimodal learning for cancer classification with missing modalities aims to build models by integrating multiple omics modalities where some modalities may be missing for some patients. Formally, let $\mathcal{D}=\{\mathbf{X}^1, \mathbf{X}^2, \dots, \mathbf{X}^M\}$ represent the $M$ omics modalities, where $\mathbf{X}^i\in\mathbb{R}^{N \times d_i}$ denotes the $i$th omics modality with $N$ patients and $d_i$ features, and let $\mathbf{y}\in\mathbb{R}^{N}$ be the corresponding labels (e.g., cancer subtypes). A binary mask $\mathbf{M}\in\{0,1\}^{N \times M}$ indicates modality availability, where $m_{ji}=1$ if the $i$th modality is available for the $j$th patient, and $m_{ji}=0$ otherwise. The objective is to learn a mapping $f_{\boldsymbol{\mathrm{\theta}}}:\mathcal{D}\to\mathbf{y}$ that handles missing modalities while learning relationships across available modalities. The model parameters $\boldsymbol{\mathrm{\theta}}$ are optimized by minimizing the loss $\mathcal{L}(\mathbf{y}, \hat{\mathbf{y}})$, where $\hat{\mathbf{y}}=f_{\boldsymbol{\mathrm{\theta}}}(\mathcal{D})$ denotes predicted labels. A summary of common notations is provided in Appendix~\ref{app:not}.

\subsection{\textbf{The \normalfont\texttt{MAGNET}} framework}

\begin{figure}
  \centering
  \includegraphics[width=0.95\linewidth]{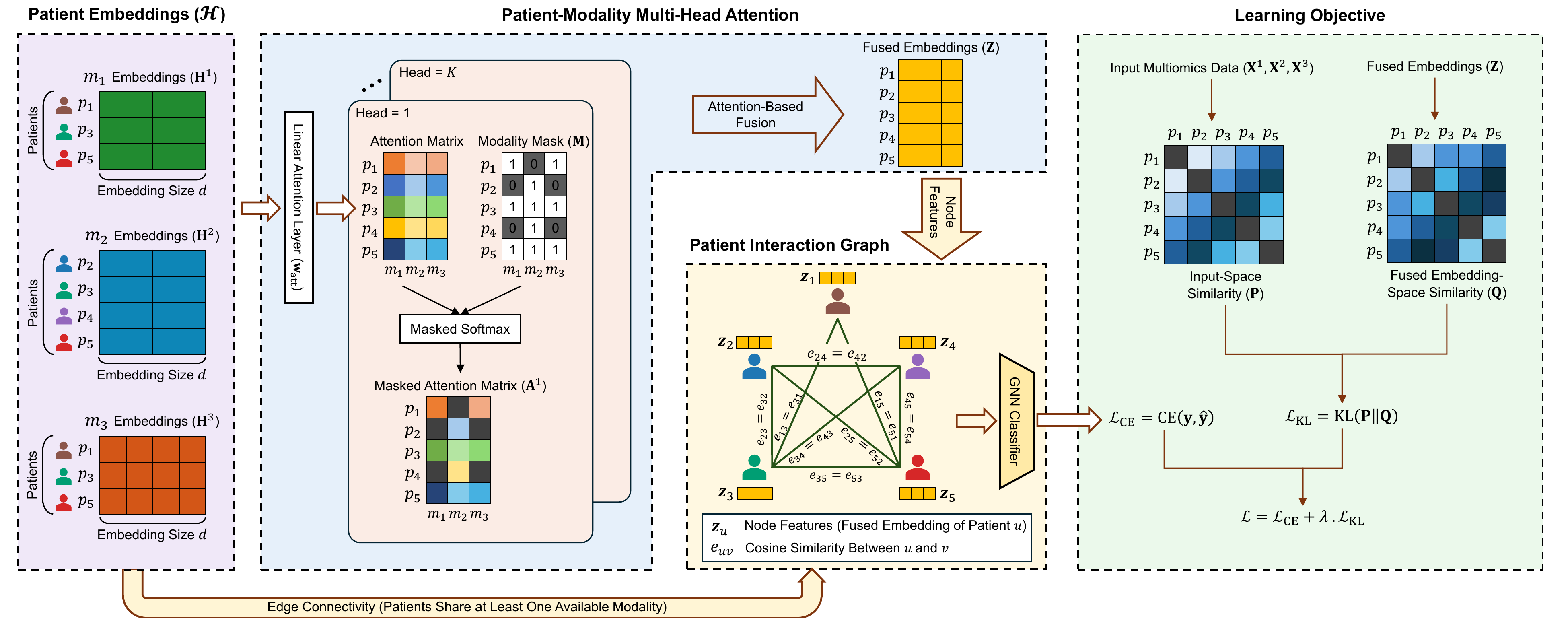}
  \caption{\texttt{MAGNET} uses a \textbf{patient-modality multi-head attention} mechanism with learnable parameters ($\mathbf{w}_{\text{att}}$) over patient embeddings ($\mathcal{H}$) and a modality mask ($\mathbf{M}$) to compute patient-specific modality attention weights ($\mathbf{A}^1, \ldots, \mathbf{A}^K$). These weights are used to aggregate patient embeddings into a fused embedding ($\mathbf{Z}$). A \textbf{patient interaction graph} is then constructed, where nodes represent patients with fused embeddings ($\mathbf{z}_u$) as node features, and edges connect patients sharing at least one available modality, with cosine similarity used as the edge feature ($e_{uv}$). A GNN learns from this graph to perform prediction. The model is optimized using cross-entropy loss ($\mathcal{L}_{\text{CE}}$) for classification and KL-divergence loss ($\mathcal{L}_{\text{KL}}$) to align the similarity distribution of the input space ($\mathbf{P}$) with the fused embedding space ($\mathbf{Q}$).}
  \label{fig:main}
\end{figure}

Figure~\ref{fig:main} illustrates the overall architecture of \texttt{MAGNET}. Conceptually, \texttt{MAGNET} consists of three modules. The first module, \textit{Modality-Specific Encoder}, encodes each modality into a lower-dimensional representation. The second module, \textit{Patient-Modality Multi-Head Attention Integration}, assigns different weights to each modality for each patient using a binary modality mask and fuses all patient embeddings across modalities into a unified embedding. The third module, \textit{Patient Graph Construction for GNN Classification}, constructs a graph to represent interactions between patients while accounting for missing modalities, on which a GNN makes the final prediction. Each module, along with the training strategy, is detailed in the following sections. 

\subsection{Modality-specific encoder}
\label{sec:encoder}

To handle the varying high dimensionality of omics modalities, modality-specific multi-layer perceptron (MLP) encoders transform each modality $\mathbf{X}^i$ into a lower-dimensional embedding  $\mathbf{H}^i\in\mathbb{R}^{N \times d}$. This yields a fixed embedding dimension $d$ across all modalities to facilitate downstream integration.

\subsection{Patient-modality multi-head attention integration}

To address missing modalities and enable the integration of patient embeddings across available modalities, we introduce a patient-modality multi-head attention (PMMHA) mechanism. This mechanism calculates attention weights for each modality specific to each patient, enabling selective weighting of modality contributions based on their importance and availability.

In a patient-modality single-head attention (PMSHA) mechanism, the modality-specific embeddings are first stacked to form the tensor $\mathcal{H}\in\mathbb{R}^{N \times M \times d}$. A linear transformation is then applied to transform this tensor into higher-level features using a learnable weight matrix $\mathbf{W}_{\text{lin}}\in\mathbb{R}^{d \times d}$.

Next, an attention coefficient matrix is calculated to evaluate the importance of each modality for each patient. To ensure comparability across modalities, the coefficients are normalized using a \textit{masked softmax} operation based on the binary modality mask $\mathbf{M}$. The process is defined as:
\begin{equation}
\label{eq:att-mha}
\mathbf{A} = \text{MaskedSoftmax}(\text{Linear}(\mathcal{H}; \mathbf{w}_{\text{att}}), \mathbf{M}),
\end{equation}
where $\mathbf{A}\in\mathbb{R}^{N \times M}$ represents the attention coefficient matrix, and $\mathbf{w}_{\text{att}} \in \mathbb{R}^{d}$ is a learnable parameter. Here, $\text{MaskedSoftmax}(\cdot, \mathbf{M})$ applies a softmax over modalities while assigning $-\infty$ to entries corresponding to missing modalities indicated by the binary mask $\mathbf{M}$, ensuring that missing modalities receive zero attention weight prior to normalization. Finally, the embeddings for each patient are fused using a weighted sum, where the attention coefficients determine the contribution of each modality. This is performed as:
\begin{equation}
\mathbf{Z}=\sum_{i=1}^{M}{\mathbf{a}^{i} \odot \mathbf{H}^{i}},
\end{equation}
where $\mathbf{Z}\in\mathbb{R}^{N \times d}$ is the fused embedding for all patients, and $\mathbf{a}^{i}\in\mathbb{R}^{N \times 1}$ is the attention weights for modality $i$ extracted from $\mathbf{A}$.

To enhance the model's ability to capture diverse patterns, PMMHA allows each head to focus on different aspects of the modalities. Extending PMSHA, the transformed embedding $\mathcal{H}$ is divided across $K$ attention heads, and the attention process is applied independently to each head. Specifically, the embedding dimension for each head is set to $d_h=d/K$, and $\mathcal{H}$ is reshaped into $\mathbb{R}^{N \times M \times d_h \times K}$. For each head $k$, the normalized attention coefficients $\mathbf{A}^k$ are calculated over the corresponding $d_h$-dimensional modality embeddings $\{\mathbf{H}_k^i\}_{i=1}^{M}$ using a head-specific learnable vector $ \mathbf{w}_{\text{att}}^k \in \mathbb{R}^{d_h}$, following the attention formulation in Equation~\ref{eq:att-mha}. These coefficients are then used to fuse the embeddings for that head to produce $\mathbf{Z}^k$. Finally, the fused embeddings from all heads are concatenated and linearly projected back to the original dimension to yield the final fused embedding $\mathbf{Z}\in\mathbb{R}^{N \times d}$.

\subsection{Patient graph construction for GNN classification}

With the fused embeddings generated for patients, we aim to make predictions by leveraging both relationships from the input multiomics data and holistic patterns captured in the fused embeddings.

\texttt{MAGNET} constructs a patient interaction graph $\mathcal{G}=(\mathcal{V},\mathcal{E})$, where $\mathcal{V}$ represents a set of $N$ nodes corresponding to patients, and $\mathcal{E}$ represents edges connecting patients. Edges are formed using the binary modality mask $\mathbf{M}$, where an edge exists between two patients if they share at least one available modality. The fused embeddings serve as the initial node features of the graph. Moreover, the cosine similarity between patients is assigned as the edge feature, computed from the input multiomics data using only shared modalities. To focus the graph on the most relevant patient correlations, only edges with cosine similarity exceeding a threshold $\beta$ are retained, and any isolated nodes are connected to their most similar neighbor to preserve graph connectivity.

Given the constructed graph $\mathcal{G}$, \texttt{MAGNET} employs a multi-layer GNN, utilizing the graph sample and aggregation (GraphSAGE) operator \cite{hamilton2017inductive}, to encode the graph. At each layer, the GNN updates a patient’s representation by aggregating feature information from its neighbors, allowing the model to incorporate local structural context and capture relationships between patients. This design also enables generating embeddings for unseen patients, as long as their neighborhood information is available \cite{hamilton2017inductive, hamilton2020graph}. The GNN stacks multiple GraphSAGE layers, with each layer $l$ defined as:
\begin{equation}
\label{eq:sage}
\mathbf{z}_u^{(l+1)} =
\text{ReLU}\left(
\mathbf{W}^{(l)}_{\text{root}} \cdot \mathbf{z}_u^{(l)} +
\mathbf{W}^{(l)}_{\text{agg}}
\cdot
\text{Mean}\left(
\left\{
\mathbf{W}^{(l)}_{\text{msg}} \cdot [\mathbf{z}_v^{(l)} \,\|\, e_{uv}]
\mid v \in \mathcal{N}(u)
\right\}
\right)
\right),
\end{equation}
where $\mathbf{z}_u^{(l)}$ is a row of $\mathbf{Z}^{(l)}$, representing the embedding of node $u$, $e_{uv}$ is the edge feature between node $u$ and a neighbor node $v$, $\mathcal{N}(u)$ is the set of directly connected neighbors of node $u$, $\,\|\,$ is the concatenation operator, and $\mathbf{W}^{{(l)}}_\text{root}$, $\mathbf{W}^{{(l)}}_\text{agg}$, and $\mathbf{W}^{{(l)}}_\text{msg}$ are learnable weight matrices. The initial node representations are the fused embeddings learned from the previous module, given by $\mathbf{Z}^{(0)}=\mathbf{Z}$. 

Finally, we use the embedding generated at the last layer $L$, $\mathbf{Z}^{(L)}$, as input to an MLP decoder to produce the final predictions, defined as $\hat{\mathbf{y}} = \text{MLP}(\mathbf{Z}^{(L)}; \mathbf{W}_{\text{MLP}})$.

\subsection{Learning objective}

We optimize the model parameters by minimizing a combined loss that balances prediction accuracy and representation quality. A \textit{supervised cross-entropy loss} is used for patient classification. To preserve patient-level relationship structure in the fused embedding space, we further use a \textit{KL divergence-based loss} designed to maintain consistency between patient similarities in the input and fused embedding spaces despite missing modalities. The supervised cross-entropy loss is defined as: 
\begin{equation}
\label{eq:loss_cls}
\mathcal{L}_{\text{CE}}=\text{CE}(\mathbf{y}, \hat{\mathbf{y}})=\sum_{j=1}^{N}{{\text{CE}}(y_j,{\hat{y}}_j)},
\end{equation}
where $\text{CE}(\cdot,\cdot)$ is the cross-entropy loss function between true label $y_j$ and predicted label $\hat{y}_j$ for pateint $j$. To preserve patient-level similarity structure in the fused embedding space, we use a KL divergence-based loss:
\begin{equation}
\label{eq:kl_loss}
\mathcal{L}_{\text{KL}} = \text{KL}(\mathbf{P} \parallel \mathbf{Q}) = \sum_{i=1}^{N} \sum_{\substack{j=1 \\ j \ne i}}^{N} p_{ij} \log \frac{p_{ij}}{q_{ij}},
\end{equation}
where $p_{ij}$ (element of $\mathbf{P}$) is the normalized cosine similarity between patients $i$ and $j$ in the input space, and $q_{ij}$ (element of $\mathbf{Q}$) is the corresponding normalized similarity in the fused embedding space, computed using a Student-$t$ kernel \cite{van2008visualizing} as:
\begin{equation}
q_{ij} = \frac{\left(1 + \|\mathbf{z}_i - \mathbf{z}_j\|^2 \right)^{-1}}{\sum_{k \ne l} \left(1 + \|\mathbf{z}_k - \mathbf{z}_l\|^2 \right)^{-1}}, 
\end{equation}
where $\|\cdot\|^2$ denotes the squared Euclidean distance between the fused embeddings of two patients. The total loss is $\mathcal{L} = \mathcal{L}_{\text{CE}} + \lambda\, \mathcal{L}_{\text{KL}}$, where $\lambda$ is a balancing coefficient.

\subsection{Training process}
\label{sec:alg}

We train \texttt{MAGNET} using an inductive learning approach, ensuring that patients in the test set are not utilized during the training phase. When constructing the patient graph, all edges connecting patients in the training set to those in the test set are excluded. Moreover, loss computation is performed exclusively on the patients in the training set. The \texttt{MAGNET} algorithm is described in Appendix~\ref{app:alg}.

\section{Experiments}
\label{sec:experiments}

\subsection{Experimental settings}
\label{sec:setting}
\paragraph{Real-world multiomics datasets} To evaluate \texttt{MAGNET}, we use three publicly available multiomics datasets from The Cancer Genome Atlas \cite{Weinstein2013tcga}, which present real-world missingness at varying rates:

\begin{itemize}
\item \textbf{Breast invasive carcinoma (BRCA):} This dataset uses the PAM50 classifier, a 50-gene signature, to categorize BRCA into five subtypes based on gene expression: Luminal A, Luminal B, basal-like, HER2-enriched (HER2), and normal-like \cite{raj2019pca}.

\item \textbf{Bladder urothelial carcinoma (BLCA):} This dataset includes bladder cancer cases, classified as low-grade or high-grade for grade classification \cite{cancer2014comprehensive}.

\item \textbf{Ovarian serous cystadenocarcinoma (OV):} 
This dataset contains ovarian cancer samples, divided into long-term (survival time $\geq 3$ years) and short-term (survival time $<3$ years with `DECEASED' status) survivors \cite{el2018min}. 
\end{itemize}

We analyze three omics modalities across all datasets: DNA methylation (DNA), RNAseq gene expression (mRNA), and miRNA  expression (miRNA). Each omics modality is individually preprocessed to remove noise and irrelevant features, with details provided in Appendix~\ref{app:data-prep-real}.

\paragraph{Simulated multiomics dataset} While our primary evaluation focuses on real-world multiomics datasets with naturally occurring missingness, we additionally evaluate \texttt{MAGNET} on a publicly available simulated multiomics cancer dataset generated using the InterSim CRAN package \cite{ma_2024_13989262}. The dataset comprises 500 samples across 15 clusters of varying sizes and three omics modalities: DNA methylation, mRNA expression, and protein expression. Further details are provided in Appendix~\ref{app:data-prep-sim}.

\paragraph{Simulated scalability dataset} To evaluate scalability with increasing numbers of modalities, we generate a controlled simulated dataset with 500 samples and varying numbers of modalities ($M=2$ to $10$). Each modality contains 1000 features, ensuring consistent modality dimensionality across settings. This dataset is used only for scalability analysis and does not represent biological data.

\paragraph{Evaluation metrics} We evaluate the performance of fusion methods on the BLCA and OV datasets for binary classification using accuracy, area under the precision-recall curve (AUPRC), area under the receiver operating characteristic curve (AUROC), and Matthews correlation coefficient (MCC). For multi-class classification on the BRCA dataset, we assess performance using accuracy, macro-averaged F1 score (Macro F1), weighted-averaged F1 score (Weighted F1), and MCC.

\paragraph{Evaluation strategies} We divide each dataset into matched (patients with all modalities) and unmatched (patients with missing modalities) data. Each subset is split into training, validation, and test sets (7:1:2), ensuring diverse missing patterns in each set. Matched and unmatched subsets are then combined to form the final sets. For hyperparameter tuning, models are trained on the training set, with the best parameters selected based on validation performance. Due to limited sample size, the training and validation sets are combined after tuning for final training. All methods are trained on this training set and evaluated on the test set. To ensure robustness, all experiments are performed five times, reporting mean and standard deviation. The same splits are used across all methods.

\paragraph{Implementation details} We implement \texttt{MAGNET} in Python 3.10 using PyTorch 2.1.0 and PyTorch Geometric 2.4.0. The Adam optimizer \cite{kingma2015adam} is used for training, with a step decay learning rate scheduler that reduces the learning rate by a factor of 0.8 every 20 epochs. All models are trained for 200 epochs, with the batch size set to the total number of patients in each dataset due to the limited number of patients. All experiments are run on an Ubuntu 24.04 machine with an NVIDIA GeForce RTX 4090 GPU. More implementation details, including hyperparameter options and the selected configurations, are provided in Appendix~\ref{app:hyper}. The source code of \texttt{MAGNET} is publicly available.\footnote{\url{https://github.com/SinaTabakhi/MAGNET}}

\paragraph{Baselines} We compare \texttt{MAGNET} with five state-of-the-art multimodal fusion methods. \textbf{MUSE} \cite{wu2024multimodal} is a recent method that leverages bipartite graphs for direct prediction. \textbf{MRGCN} \cite{yang2023mrgcn} uses an encoder-decoder framework based on graph convolutional networks (GCNs) for direct predictions with missing modalities. \textbf{M3Care} \cite{zhang2022m3care} is an imputation-based method that reconstructs missing modalities in the latent space by leveraging similarity between patients. MOGONET \cite{wang2021mogonet} is a supervised multiomics integration method based on GCN that requires complete modalities. To address missing modalities in MOGONET, we apply two imputation strategies, zero imputation and $k$-nearest neighbor ($k$NN), referred to as \textbf{MOGONET-Zero} and \textbf{MOGONET-$k$NN}. More details on these baselines are provided in Appendix~\ref{app:baseline}.

\subsection{Performance of cancer classification}

Table~\ref{tbl:cls} presents the classification performance of fusion methods on the BRCA, BLCA, and OV datasets. \texttt{MAGNET} consistently outperforms baseline multimodal fusion methods across all datasets and evaluation metrics, except for AUROC on OV, where it achieves the second-best result. 

\begin{table}
\caption{ \textbf{Real-world} classification performance comparison on three datasets: BRCA, BLCA, and OV.  Reported values are the mean ± standard deviation over five independent runs (\textbf{best}, \underline{second-best}).}
\label{tbl:cls}
\setlength{\tabcolsep}{3.5pt}
\centering
\begin{scriptsize}
\begin{tabular}{llcccccc}
\toprule
Dataset & Metric & MOGONET-Zero \cite{wang2021mogonet} & MOGONET-$k$NN \cite{wang2021mogonet} & MRGCN \cite{yang2023mrgcn} & M3Care \cite{zhang2022m3care} & MUSE \cite{wu2024multimodal} & \texttt{MAGNET} \\
\midrule
BRCA
& Accuracy (\textbf{\textcolor{teal}{\textuparrow}}) & 0.795{\tiny$\pm$0.025} & 0.847{\tiny$\pm$0.015} & 0.844{\tiny$\pm$0.009} & \underline{0.899{\tiny$\pm$0.016}} & 0.895{\tiny$\pm$0.021} & \textbf{0.918{\tiny$\pm$0.012}} \\
& Macro F1 (\textbf{\textcolor{teal}{\textuparrow}}) & 0.638{\tiny$\pm$0.062} & 0.794{\tiny$\pm$0.035} & 0.826{\tiny$\pm$0.014} & 0.872{\tiny$\pm$0.018} & \underline{0.880{\tiny$\pm$0.014}} & \textbf{0.902{\tiny$\pm$0.019}} \\
& Weighted F1 (\textbf{\textcolor{teal}{\textuparrow}}) & 0.758{\tiny$\pm$0.036} & 0.838{\tiny$\pm$0.018} & 0.842{\tiny$\pm$0.008} & \underline{0.898{\tiny$\pm$0.015}} & 0.894{\tiny$\pm$0.022} & \textbf{0.917{\tiny$\pm$0.011}} \\
& MCC (\textbf{\textcolor{teal}{\textuparrow}}) & 0.706{\tiny$\pm$0.035} & 0.781{\tiny$\pm$0.022} & 0.778{\tiny$\pm$0.012} & \underline{0.858{\tiny$\pm$0.022}} & 0.851{\tiny$\pm$0.031} & \textbf{0.884{\tiny$\pm$0.016}} \\
\midrule
BLCA 
& Accuracy (\textbf{\textcolor{teal}{\textuparrow}}) & 0.952{\tiny$\pm$0.005} & 0.952{\tiny$\pm$0.005} & 0.955{\tiny$\pm$0.000} & \underline{0.968{\tiny$\pm$0.011}} & 0.966{\tiny$\pm$0.014} & \textbf{0.970{\tiny$\pm$0.006}} \\
& AUPRC (\textbf{\textcolor{teal}{\textuparrow}}) & 0.652{\tiny$\pm$0.106} & \underline{0.688{\tiny$\pm$0.098}} & 0.617{\tiny$\pm$0.086} & 0.686{\tiny$\pm$0.114} & 0.653{\tiny$\pm$0.085} & \textbf{0.724{\tiny$\pm$0.101}} \\
& AUROC (\textbf{\textcolor{teal}{\textuparrow}}) & 0.898{\tiny$\pm$0.142} & 0.902{\tiny$\pm$0.162} & 0.944{\tiny$\pm$0.033} & \underline{0.949{\tiny$\pm$0.051}} & 0.939{\tiny$\pm$0.046} & \textbf{0.956{\tiny$\pm$0.025}} \\
& MCC (\textbf{\textcolor{teal}{\textuparrow}}) & -0.005{\tiny$\pm$0.009} & -0.005{\tiny$\pm$0.009} & 0.000{\tiny$\pm$0.000} & \underline{0.597{\tiny$\pm$0.164}} & 0.509{\tiny$\pm$0.294} & \textbf{0.642{\tiny$\pm$0.072}} \\
\midrule
OV 
& Accuracy (\textbf{\textcolor{teal}{\textuparrow}}) & 0.581{\tiny$\pm$0.027} & 0.573{\tiny$\pm$0.051} & 0.597{\tiny$\pm$0.014} & 0.597{\tiny$\pm$0.031} & \underline{0.608{\tiny$\pm$0.036}} & \textbf{0.614{\tiny$\pm$0.052}} \\
& AUPRC (\textbf{\textcolor{teal}{\textuparrow}}) & \underline{0.630{\tiny$\pm$0.030}} & 0.594{\tiny$\pm$0.044} & \textbf{0.646{\tiny$\pm$0.022}} & 0.607{\tiny$\pm$0.051} & 0.628{\tiny$\pm$0.078} & \textbf{0.646{\tiny$\pm$0.046}} \\
& AUROC (\textbf{\textcolor{teal}{\textuparrow}}) & 0.621{\tiny$\pm$0.037} & 0.603{\tiny$\pm$0.062} & \textbf{0.655{\tiny$\pm$0.025}} & 0.630{\tiny$\pm$0.040} & \underline{0.652{\tiny$\pm$0.067}} & \underline{0.652{\tiny$\pm$0.056}} \\
& MCC (\textbf{\textcolor{teal}{\textuparrow}}) & 0.199{\tiny$\pm$0.071} & 0.126{\tiny$\pm$0.138} & 0.201{\tiny$\pm$0.029} & 0.199{\tiny$\pm$0.060} & \underline{0.225{\tiny$\pm$0.073}} & \textbf{0.228{\tiny$\pm$0.104}} \\
\bottomrule
\end{tabular}
\end{scriptsize}
\end{table}

\paragraph{Results on BRCA} \texttt{MAGNET} outperforms the best-performing direct prediction counterparts, MRGCN and MUSE, with improvements of 2.3\% in accuracy, 2.2\% in Macro F1, 2.3\% in Weighted F1, and 3.88\% in MCC. MUSE underperforms likely due to overlapping feature distributions among certain BRCA classes, as it does not explicitly model direct patient-to-patient connections. Similarly, MRGCN fuses patient embeddings across modalities by assigning equal contributions, which may overlook varying modality importance. Moreover, both methods may struggle to handle severe modality missingness, as their design assumptions might not effectively capture complex missing patterns. This limitation is further reflected in the performance of MOGONET-Zero, which performs worst across all metrics, suggesting that simple zero imputation is ineffective under such missingness.

\paragraph{Results on BLCA} \texttt{MAGNET} maintains its superior performance, achieving a 3.6\% improvement in AUPRC and a 7.54\% relative improvement in MCC over the best baseline, a particularly notable gain given the dataset's class imbalance. In contrast, MOGONET with imputation yields the worst results across nearly all metrics for BLCA. This may be due to BLCA’s relatively low rate of missing modalities combined with a highly imbalanced class distribution, where even minor imputations can introduce errors that especially affect the already limited minority class.

\paragraph{Results on OV} On this well-balanced dataset, \texttt{MAGNET} demonstrates strong overall classification performance, achieving the highest accuracy, AUPRC, and MCC. However, its focus on correctly identifying the positive class may slightly affect ranking performance across all thresholds, leading to a marginally lower AUROC. Despite this, the overall results confirm the robustness of the method. Notably, while M3Care performs well on BRCA and BLCA, it ranks among the worst on OV, highlighting the difficulty for baselines to generalize across datasets.

\texttt{MAGNET}’s superior results highlight the effectiveness of its PMMHA mechanism and graph-based modeling, enhancing multimodal fusion in the presence of missing modalities. 

\subsection{Simulated missing-modality studies}

\begin{figure}
  \centering
  \includegraphics[width=0.95\linewidth]{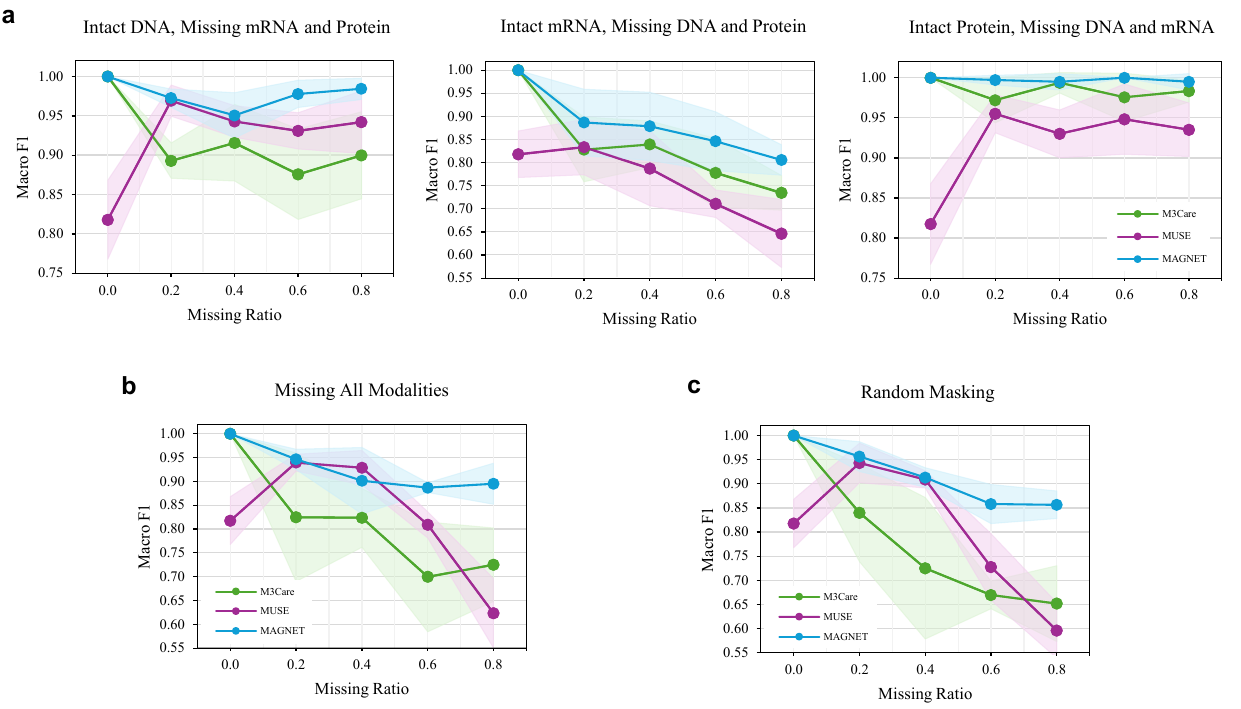}
    \caption{Effect of varying missingness ratios on simulated multiomics dataset using Macro F1, averaged over five independent runs per ratio. (a) One modality remains intact while the other two are uniformly subsampled. (b) A subset of patients is shared across all modalities, with the rest uniquely assigned to individual modalities. (c) Modalities are randomly masked with different probabilities.}
  \label{fig:synth}
\end{figure}

We compare \texttt{MAGNET} with the top-performing baselines, M3Care and MUSE, on the simulated multiomics dataset under several missing-modality scenarios, with results shown in Figure~\ref{fig:synth}.

In the first scenario, we keep one omics modality intact while uniformly subsampling the other two at missingness ratios ranging from 0.2 to 0.8, with zero missingness representing the complete dataset (Figure~\ref{fig:synth}a). The results show that \texttt{MAGNET} consistently outperforms the baselines across all missingness levels. Moreover, when DNA or protein modalities are kept intact, performance remains relatively stable, indicating that they are more informative. In contrast, when mRNA is the only intact modality, performance drops significantly, suggesting it is less predictive on its own.

In the second scenario, we retain a subset of patients that are shared across all three modalities and evenly assign the remaining patients to individual modalities. The missingness ratio varies from 0.2 to 0.8 (Figure~\ref{fig:synth}b). We observe that \texttt{MAGNET} consistently achieves the highest performance across most settings. Notably, as the missingness ratio increases, the performance gap between \texttt{MAGNET} and the baseline methods also grows, highlighting the model’s robustness to missing modality information.

In the third scenario, we adopt a more complex setup where no omics modality is kept fully intact. Modalities are randomly masked with probabilities ranging from 0.2 to 0.8 (Figure~\ref{fig:synth}c). Performance results show that \texttt{MAGNET} consistently outperforms the baselines across all missingness levels. We also observe that as the masking probability increases, the performance gap between \texttt{MAGNET} and the baselines gets larger. For example, under severe missingness (80\%), \texttt{MAGNET} improves performance by around 20\% and 26\% compared to M3Care and MUSE, respectively, demonstrating strong resilience to high levels of missing modalities. We also observe that M3Care, as an imputation-based method, experiences a much larger performance drop than the other methods as the missing probability increases. This may be because imputation-based methods rely on reconstructing missing data, which becomes increasingly difficult and error-prone as more information is missing.

\subsection{Scalability analysis}
\begin{wrapfigure}{r}{0.4\linewidth}
  \vspace{-14pt}
  \centering
  \includegraphics[width=0.9\linewidth]{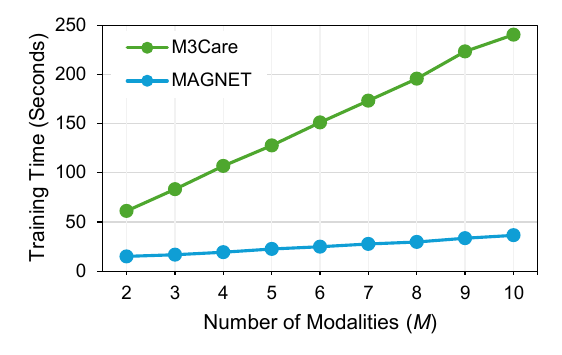}
  \vspace{-8pt}
  \caption{Training time vs. number of modalities on a simulated dataset.}
  \label{fig:modality_scalability}
  \vspace{-6pt}
\end{wrapfigure}

We evaluate scalability with increasing modalities ($M=2$ to $10$) on the simulated scalability dataset (Figure~\ref{fig:modality_scalability}). Missing modalities are simulated through random masking with probability 0.5, and results are averaged over five independent runs. The results show that \texttt{MAGNET} exhibits approximately linear growth in training time while maintaining a substantial efficiency advantage over M3Care. In contrast, the training time of M3Care increases much more rapidly as modalities increase. At $M=2$, \texttt{MAGNET} is already 4.1$\times$ faster than M3Care; by $M=10$, this gap widens significantly, with \texttt{MAGNET} achieving a 6.6$\times$ training-time improvement. These results show that \texttt{MAGNET} scales efficiently with the number of modalities.

\subsection{Ablation study}

\begin{wraptable}{r}{0.54\linewidth}
  \vspace{-16pt}
  \caption{Ablation study on the impact of each individual component of \texttt{MAGNET} (\textbf{best}, \underline{second-best}).}
  \vspace{6pt}
  \label{tbl:abl-cls}
  \setlength{\tabcolsep}{2.5pt}
  \centering
  \begin{scriptsize}
  \begin{tabular}{lllll}
    \toprule
    ID &  Method & BRCA & BLCA & OV \\
    \midrule
    A1 & \texttt{MAGNET} w/o PMMHA        & 0.905{\tiny$\pm$0.016} & 0.681{\tiny$\pm$0.064} & \underline{0.600{\tiny$\pm$0.049}} \\
    A2 & \texttt{MAGNET} w/o GNN          & 0.907{\tiny$\pm$0.019} & 0.675{\tiny$\pm$0.068} & \underline{0.600{\tiny$\pm$0.041}} \\
    A3 & \texttt{MAGNET} w/o Edge Feature & 0.910{\tiny$\pm$0.015} & \underline{0.719{\tiny$\pm$0.135}} & \underline{0.600{\tiny$\pm$0.064}} \\
    A4 & \texttt{MAGNET} w/o KL Loss      & \underline{0.911{\tiny$\pm$0.013}} & 0.686{\tiny$\pm$0.104} & 0.592{\tiny$\pm$0.071} \\
    A5 & \texttt{MAGNET} w/ GAT           & 0.711{\tiny$\pm$0.083} & 0.606{\tiny$\pm$0.127} & 0.551{\tiny$\pm$0.029} \\
    A6 & \texttt{MAGNET} w/ GCN           & 0.823{\tiny$\pm$0.002} & 0.609{\tiny$\pm$0.110} & 0.567{\tiny$\pm$0.041} \\
    A7 & \texttt{MAGNET} w/ GIN           & 0.560{\tiny$\pm$0.173} & 0.490{\tiny$\pm$0.157} & 0.529{\tiny$\pm$0.076} \\
    \midrule 
    -  & \texttt{MAGNET}                  & \textbf{0.918{\tiny$\pm$0.012}} & \textbf{0.724{\tiny$\pm$0.101}} & \textbf{0.614{\tiny$\pm$0.052}} \\
    \bottomrule
  \end{tabular}
  \end{scriptsize}
   \vspace{-10pt}
\end{wraptable}

We conduct an ablation study to evaluate the contribution of individual components in \texttt{MAGNET}. We consider seven modifications: (A1) removing PMMHA and assigning equal modality contributions, (A2) removing the GNN and applying the MLP decoder to the fused embedding, (A3) removing edge features from the patient graph, (A4) removing the KL loss and using only the cross-entropy loss, and (A5)-(A7) replacing GraphSAGE with GAT \cite{veličković2018graph}, GCN \cite{Kipf2017SemiSupervisedCW}, and GIN \cite{xu2018how}, respectively. Table~\ref{tbl:abl-cls} presents the results. We observe a significant performance drop when PMMHA and the GNN are removed, indicating that simple aggregation is insufficient for effective fusion under missing modalities, and that graph-based interactions are essential for capturing patient relationships. This is further supported by (A5)-(A7), where replacing the original GNN leads to a consistent drop, highlighting the suitability of GraphSAGE with edge features. Overall, these results validate the effectiveness of \texttt{MAGNET}.

\section{Conclusion and limitations}
\label{sec:conc}
\paragraph{Contributions} In this paper, we propose \texttt{MAGNET}, a novel method for direct prediction using partial omics modalities without the need for imputation or patient exclusion in cancer classification. \texttt{MAGNET} introduces a patient-modality multi-head attention mechanism to fuse patient representations by learning modality importance for each patient while adaptively masking missing modalities. It also constructs a patient interaction graph using missing-modality patterns for connectivity and the learned embeddings as node features. To preserve patient-level structure during fusion, \texttt{MAGNET} further incorporates a KL divergence-based loss that aligns similarity distributions before and after fusion. Key strengths of \texttt{MAGNET} include its linear complexity, extensibility to new modalities, and robustness to diverse missing-modality patterns in both training and test data. Experimental results on three multiomics datasets with real-world missingness demonstrate its superiority over baseline methods. Furthermore, quantitative analysis of the learned representations further validates its effectiveness.

\paragraph{Limitations} This work focuses on biological datasets, specifically multiomics data for cancer classification, allowing targeted evaluation in a clinically relevant context. Future work will explore the integration of fundamentally different modalities, such as imaging and text, to support broader applicability. Although our experiments are limited to cancer-related applications, the proposed method is general and has the potential to be applied to non-biomedical domains.

\paragraph{Broader impact} This work advances machine learning methods for multimodal biomedical data, supporting cancer research in settings with missing modalities. By enabling adaptive fusion under missing modalities, it may improve downstream analyses in precision medicine. While the method is not intended for direct clinical deployment, careful validation would be required before clinical use.

\medskip

\bibliography{main}
\bibliographystyle{plain}

\newpage
\appendix

\section{Notations}
\label{app:not}

The common notations used in this paper are provided in Table~\ref{tbl:not}.

\begin{table}[h]
  \caption{Notations and descriptions.}
  \label{tbl:not}
  \centering
    \begin{tabular}{ll}
      \toprule
      Notation & Description \\
      \midrule
        $\mathbf{a}^i \in \mathbb{R}^{N \times 1}$   & Attention weights for modality $i$ extracted from $\mathbf{A}$ \\
        $\mathbf{A} \in \mathbb{R}^{N \times M}$   & Attention coefficient matrix in the PMMHA \\
        $\beta$   & Edge discard threshold in graph $\mathcal{G}$ \\
        $d_h$   & Embedding size for each head in the PMMHA  \\
        $\mathcal{D}=\{\mathbf{X}^1, \mathbf{X}^2, \dots, \mathbf{X}^M\}$   & Multiomics dataset with $M$ modalities \\
        $e_{uv}$   & Edge feature between nodes $u$ and $v$ in graph $\mathcal{G}$ \\
        $\mathcal{G}=(\mathcal{V},\mathcal{E})$   & Patient interaction graph with nodes $\mathcal{V}$ and edges $\mathcal{E}$ \\
        $\mathcal{H} \in \mathbb{R}^{N \times M \times d}$   & Stacked representations across modalities \\
        $\mathbf{H}^i \in \mathbb{R}^{N \times d}$      & Representation of the $i$th modality of dimension $d$ \\
        $K$   & Number of heads in the PMMHA \\
        $L$   & Number of layers in the GNN \\
        $\lambda$   & Balancing coefficient between the cross-entropy and KL divergence losses \\
        $\mathbf{M} \in \{0,1\}^{N \times M}$      & Binary modality mask matrix \\
        $M$    & Number of omics modalities \\
        $\mathcal{N}(u)$   & Neighbors of node $u$ in graph $\mathcal{G}$\\
        $N$     & Number of patients \\
        $\mathbf{X}^i \in \mathbb{R}^{N \times d_i}$ & $i$th omics modality with $N$ patients and $d_i$ features \\
        $\mathbf{y}, \hat{\mathbf{y}} \in \mathbb{R}^N$ & True and predicted labels for $N$ patients \\
        $\mathbf{z}_u^{(l)}$   & Embedding of node $u$ at the $l$th layer of GNN \\
        $\mathbf{Z} \in \mathbb{R}^{N \times d}$   & Fused embeddings for all patients \\
      \bottomrule
    \end{tabular}
\end{table}

\section{Algorithm for \normalfont\texttt{MAGNET}}
\label{app:alg}
Algorithm~\ref{alg:magnet} outlines the \texttt{MAGNET} strategy for direct prediction with partial modalities.

\begin{algorithm}[h]
    \caption{\texttt{MAGNET} Training}
    \label{alg:magnet}
    {\bfseries Input:} Training multiomics dataset $\mathcal{D}$, binary modality mask $\mathbf{M}$, number of training epochs $T$\\
    {\bfseries Output:} Predicted labels ${\hat{\mathbf{y}}\in\mathbb{R}^N}$
    \begin{algorithmic}[1]
    \STATE Construct patient interaction graph $\mathcal{G}$ with edges formed using $\mathbf{M}$
    \STATE Set edge features in $\mathcal{G}$ using cosine similarity between patients computed from $\mathcal{D}$
    \FOR{$t=1$ {\bfseries to} $T$}
        \FOR{$i = 1$ {\bfseries to} $M$}
            \STATE Encode modality $i$ using a modality-specific MLP
        \ENDFOR     
        \STATE Compute attention coefficients $\mathbf{A}$ for each head using Equation~\ref{eq:att-mha}
        \STATE Fuse patient embeddings to obtain $\mathbf{Z}$
        \STATE Initialize graph node features in $\mathcal{G}$ with $\mathbf{Z}$
        \STATE Apply GNN to $\mathcal{G}$ to learn patient embeddings via Equation~\ref{eq:sage}
        \STATE Predict labels $\hat{\mathbf{y}}$ using the MLP decoder
        \STATE Compute the cross-entropy loss using Equation~\ref{eq:loss_cls}
        \STATE Compute the KL divergence loss using Equation~\ref{eq:kl_loss}
        \STATE Update model parameters using the total loss
    \ENDFOR
    \end{algorithmic}
\end{algorithm}

\section{Computational complexity of \normalfont\texttt{MAGNET}}
\label{app:comp_analysis}
The computational complexity of \texttt{MAGNET} is designed to scale efficiently with the number of patients and modalities. The process consists of three main stages:

\paragraph{Modality-specific encoding} Each modality is independently mapped to a $d$-dimensional embedding via modality-specific MLPs. The computational complexity is $\mathcal{O}(N \cdot M \cdot d_i \cdot d)$, where $N$ is the number of patients, $M$ the number of modalities, and $d_i$ the input feature dimension of modality $i$. Since patients are processed independently, this stage can be efficiently parallelized and scales linearly with $N$, making it suitable for large-scale cohorts.

\paragraph{Patient-modality multi-head attention} PMMHA computes attention across modalities within each patient, without modeling interactions between patients. For a fixed total embedding dimension $d$, the multi-head design does not change the overall complexity order, since $K$ heads operate on subspaces of dimension $d/K$. Therefore, this step scales linearly with the number of patients, with complexity $\mathcal{O}(N \cdot M \cdot d)$, and avoids quadratic overhead.

\paragraph{Graph construction and GNN learning} Constructing the graph requires computing pairwise similarities between patients, resulting in a one-time cost of $\mathcal{O}(N^2)$. However, this step is performed only once as a preprocessing stage and does not affect the per-epoch training complexity. During training, scalability is determined by the number of nodes and edges in the graph. To ensure efficiency, we construct a sparse graph by retaining only the most relevant connections, which significantly reduces the number of edges compared to a fully connected graph. As a result, GNN complexity becomes proportional to the number of edges rather than $\mathcal{O}(N^2)$.

Here, we use $\mathcal{O}(N^2)$ pairwise similarity computation due to the relatively small number of patients in our datasets. However, \texttt{MAGNET} can scale to larger datasets, as the graph construction can be reduced to approximately $\mathcal{O}(N^{1.14})$ using approximate nearest neighbor methods such as NN-descent \cite{wei2011efficient} and Relative NN-descent \cite{naoki2023relative}, or to $\mathcal{O}(N \log N)$ using graph-based methods such as HNSW \cite{8594636}. Furthermore, parallelization of similarity computations across CPU/GPU resources can further reduce runtime overhead in practice.

\section{Additional details on datasets and preprocessing}
\label{app:data-prep}
\subsection{Real-world multiomics datasets} 
\label{app:data-prep-real}
All three datasets supporting the findings of this work are publicly available and obtained from the UCSC Xena platform.\footnote{\url{https://xenabrowser.net/datapages/}} For the DNA modality, we use the Illumina Infinium HumanMethylation27 for BRCA and OV, and the HumanMethylation450 for BLCA. The mRNA modality uses the Illumina HiSeq pancan normalized version, and the miRNA modality includes Illumina HiSeq data. Further details are available on the UCSC Xena platform \cite{goldman2020visualizing}.

To prepare datasets for analysis, we preprocess each omics modality separately. First, we remove features with more than 10\% missing values and fill the remaining missing values using the feature-wise mean. Second, we normalize each feature to the [0, 1] range using min-max scaling \cite{you2020handling}. Third, we exclude low-variance features with limited discriminatory ability, applying modality-specific thresholds: 0.04 for DNA in BRCA and OV, 0.08 for DNA in BLCA, and 0.03 for mRNA across all datasets. No variance filtering is applied to miRNA due to its smaller number of features. Given the high dimensionality of each omics modality, we further reduce irrelevant and redundant features using the ANOVA \cite{girden1992anova} feature selection method implemented in scikit-learn. For DNA and mRNA, we retain the top 1,000 selected features, while no feature selection is applied to miRNA due to its limited number of features. Table~\ref{tbl:data} presents an overview of the dataset statistics. The preprocessed data are available.\footnote{\url{https://github.com/SinaTabakhi/MAGNET}}

\begin{table}
  \caption{Summary of multiomics data characteristics.}
  \label{tbl:data}
  \centering
  \begin{small}
  \begin{tabular}{lccccccc}
    \toprule
        \multicolumn{1}{l}{Dataset} & \multicolumn{1}{c}{\#Patients} & \multicolumn{3}{c}{\#Patients in Omics (Missing Rate \%)} & \multicolumn{3}{c}{\#Selected Features} \\
         \cmidrule(lr){3-5} \cmidrule(lr){6-8}
        & & DNA & mRNA & miRNA & DNA & mRNA & miRNA \\
    \midrule  
        BRCA & 956 & 328 (65.69) & 956 (00.00) & 584 (38.91) & 1,000 & 1,000 & 436 \\
        BLCA & 433 & 431 (00.46) & 423 (02.31) & 426 (01.62) &  1,000 &  1,000 & 471 \\
        OV & 360 & 356 (01.11) & 184 (48.89) & 302 (16.11) &  1,000 &  1,000 & 448 \\
    \bottomrule
  \end{tabular}
  \end{small}
\end{table}

\subsection{Simulated multiomics dataset}
\label{app:data-prep-sim}
We use a publicly available simulated multiomics cancer dataset generated by the InterSim CRAN package, consisting of 500 samples across 15 clusters of varying sizes, reflecting realistic clinical scenarios. The dataset includes three omics modalities: DNA methylation, mRNA expression, and protein expression. The data is available on Zenodo \cite{ma_2024_13989262}, and further details about the data generation process can be found in the original publication \cite{ma2025moving}.

To prepare the dataset for analysis, we normalize each feature to the [0, 1] range using min-max scaling, applied separately to each omics modality. The data is then split into training and test sets with a ratio of 8:2. For all methods, we use hyperparameters similar to those used for the BRCA dataset, given their comparable nature. 

\section{Additional details on hyperparameter tuning}
\label{app:hyper}
To ensure a fair comparison, we perform hyperparameter tuning for \texttt{MAGNET} and all baseline methods using the Ray Tune library \cite{liaw2018tune}. We tune key hyperparameters that significantly influence model performance. For all methods, we run 100 trials with the Asynchronous Successive Halving Algorithm scheduler \cite{li2020system} to prioritize promising configurations. Each trial runs for up to 100 epochs, with training on the training set and evaluation on the validation set. To ensure robustness across data splits, we repeat each configuration five times and use the average performance for selection. Hyperparameter ranges are similar across methods unless otherwise specified in their original papers, in which case we adopt the recommended ranges. Table~\ref{tbl:hyp} lists the search spaces and the best-performing values for each dataset and method.

For \texttt{MAGNET}, we fix the number of layers in the MLP encoder, GNN, and MLP decoder to two across all datasets. The $\lambda$ parameter is also set to 0.1. 

\begin{table}[t]
\caption{Hyperparameter ranges and selected values across methods and datasets.}
\label{tbl:hyp}
\setlength{\tabcolsep}{5pt}
\centering
\begin{scriptsize}
\begin{tabular}{lllrrr}
\toprule
Method & Hyperparameter & Range & BRCA & BLCA & OV \\
\midrule  
\texttt{MAGNET} & MLP Hidden Dimension        & Grid Search ([128, 256])                 & 128 & 256 & 256 \\
                & Patient Graph Sparsity Rate & Discrete Choice ([0.50-0.95], step 0.05) & 0.60 & 0.60 & 0.80 \\
                & Dropout Rate                & Discrete Choice ([0.0-0.3], step 0.1)    & 0.1 & 0.2 & 0.2 \\
                & \#Heads in PMMHA            & Grid Search ([2, 4, 8])                  & 2 & 8 & 8 \\
                & Learning Rate               & Log-Uniform (0.00001, 0.001)             & 0.00032 & 0.00017 & 0.000212 \\
\midrule  
MOGONET-Zero & Hidden Dimension            & Grid Search ([128, 256])                 & 256 & 256 & 128 \\
             & \#Edges Retained per Node   & Integer Uniform ([2-10])                 & 5 & 4 & 10 \\
             & Dropout Rate                & Discrete Choice ([0.0-0.3], step 0.1)    & 0.3 & 0.3 & 0.1 \\
             & Pretraining Learning Rate   & Log-Uniform (0.00001, 0.001)             & 0.00076 & 0.00014 & 0.000033 \\
             & Training Learning Rate      & Log-Uniform (0.00001, 0.001)             & 0.00087 & 0.001 & 0.00074 \\
\midrule  
MOGONET-$k$NN & Hidden Dimension            & Grid Search ([128, 256])                 & 128 & 256 & 128 \\
              & \#Edges Retained per Node   & Integer Uniform ([2-10])                 & 2 & 2 & 2 \\
              & Dropout Rate                & Discrete Choice ([0.0-0.3], step 0.1)    & 0.0 & 0.3 & 0.2 \\
              & Pretraining Learning Rate   & Log-Uniform (0.00001, 0.001)             & 0.00095 & 0.00029 & 0.00021 \\
              & Training Learning Rate      & Log-Uniform (0.00001, 0.001)             & 0.00083 & 0.00092 & 0.0008 \\
\midrule  
MRGCN  & \#Neighbors per Node   & Discrete Choice ([5, 10, 15, 20])        & 5 & 15 & 5 \\
       & Dropout Rate           & Discrete Choice ([0.0-0.3], step 0.1)    & 0.2 & 0.0 & 0.0 \\
       & Learning Rate          & Log-Uniform (0.00001, 0.001)             & 0.00023 & 0.000024 & 0.00011 \\
\midrule  
M3Care & Hidden Dimension & Grid Search ([128, 256])                 & 256 & 256 & 128 \\
       & Dropout Rate     & Discrete Choice ([0.0-0.3], step 0.1)    & 0.2 & 0.3 & 0.1 \\
       & Learning Rate    & Log-Uniform (0.00001, 0.001)             & 0.0002 & 0.00041 & 0.00019 \\
\midrule  
MUSE & Hidden Dimension & Grid Search ([128, 256])                 & 256 & 256 & 256 \\
     & Dropout Rate     & Discrete Choice ([0.0-0.3], step 0.1)    & 0.3 & 0.1 & 0.3 \\
     & Learning Rate    & Log-Uniform (0.00001, 0.001)             & 0.00072 & 0.00018 & 0.00035 \\
\bottomrule
\end{tabular}
\end{scriptsize}
\end{table}

\section{Additional details on baselines}
\label{app:baseline}
We compare the performance of \texttt{MAGNET} with the following state-of-the-art baseline fusion methods.

\begin{itemize}
\item \textbf{MUSE \cite{wu2024multimodal}} is a recently developed method for direct prediction in the presence of missing modalities in healthcare. This framework constructs a bipartite graph with modalities and patients as nodes, where the edge features represent the patient-specific features within each modality. From this initial graph, an additional graph is generated using edge dropout. These two bipartite graphs are then processed through a Siamese graph neural network to learn patient representations, which are used for final predictions through an MLP decoder.
\item \textbf{MRGCN \cite{yang2023mrgcn}} is a direct prediction method designed to handle missing modalities in multiomics data. It follows an encoder-decoder framework, where omics-specific encoders, based on GCN, generate representations that are simply aggregated to form a shared fusion across modalities. The decoder then reconstructs the input data structure. The method originally employs graph clustering on the fused data to further optimize the algorithm and facilitate downstream tasks. To ensure a fair comparison with \texttt{MAGNET}, we replace the clustering component with a two-layer MLP and use a cross-entropy loss function, similar to the architecture used in \texttt{MAGNET}.
\item \textbf{M3Care \cite{zhang2022m3care}} is an imputation-based method for handling missing modalities in multimodal healthcare data. It first extracts latent representations from each input modality and computes patient similarities using learned deep kernels. These similarities are aggregated across available modalities to form a unified similarity matrix. Using this matrix, a patient graph is constructed for each modality by identifying similar patients, and a GNN is applied to learn patient representations. Missing modalities are imputed in the embedding space based on these learned graphs. Finally, patient representations from all modalities are fused using a Transformer-based module for downstream tasks.
\item \textbf{MOGONET \cite{wang2021mogonet}} is a supervised, late-fusion multiomics method. It employs omics-specific GCN to make initial predictions and then integrates label-level information from each omics modality using a view correlation discovery network to generate the final prediction. MOGONET assumes that all omics modalities are available for all patients. We consider MOGONET as a representative of this category and handle missing modalities using two imputation strategies: zero imputation and $k$NN imputation, referred to as \textbf{MOGONET-Zero} and \textbf{MOGONET-$k$NN}, respectively. For $k$NN-based imputation, we set $k=10$ due to the limited number of patients in the multiomics datasets.
\end{itemize}

To ensure a fair comparison, all methods are trained for 200 epochs. For MOGONET, which includes both a pretraining and training phase, we allocate 100 epochs for pretraining and 100 epochs for training. All baselines are implemented using an inductive learning approach, where test sets are excluded during the training phase.

\section{Additional experiments}
\label{app:exp}

\subsection{Importance of individual omics and their integration}
\label{app:modality_ablation}
\begin{figure}
  \centering
  \includegraphics[width=0.9\linewidth]{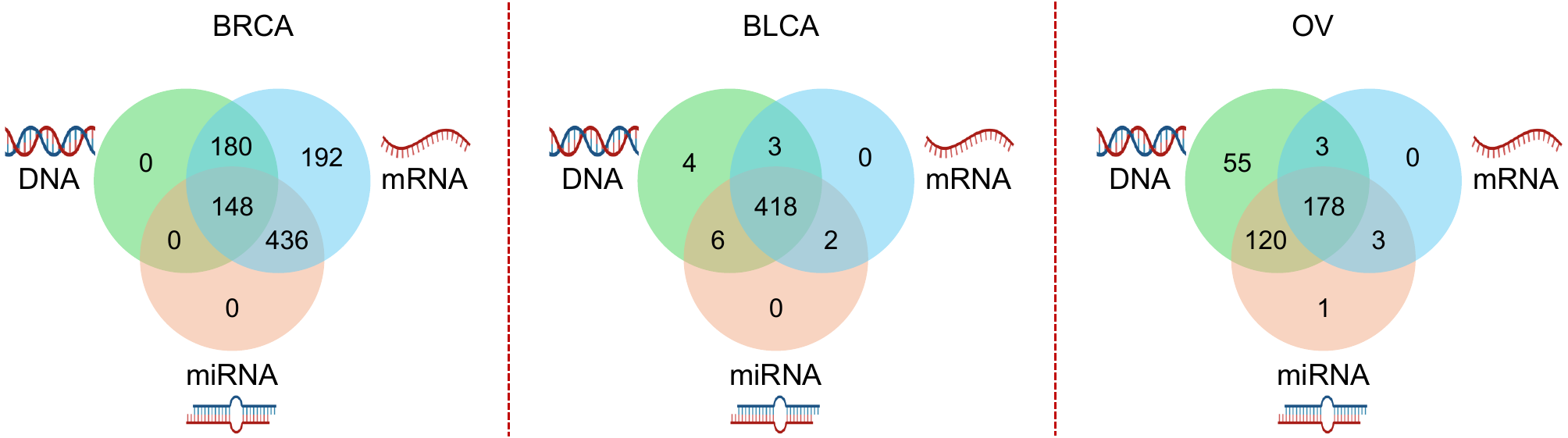}
    \caption{Statistics of three real-world multiomics datasets across three omics modalities. The Venn diagrams show the number of patients available in each modality, their pairwise overlaps, and the intersection across all three modalities (center). While single-modality methods cannot exploit complementary information across modalities, multimodal methods that require complete modalities restrict learning to a small subset of patients. \texttt{MAGNET} uses all modality combinations, enabling robust learning under diverse missing-modality patterns.}
  \label{fig:multimodal_motivation}
\end{figure}

There are two key reasons for integrating multiple omics modalities. First, relying on a single modality limits prediction for patients missing that modality. For example, as shown in Table~\ref{tbl:data}, the BRCA dataset has approximately 66\% missing DNA data. Using only DNA would exclude over half of the patients, highlighting the need to leverage other available modalities. Second, additional modalities can provide complementary information that enhances model performance beyond what a single modality can achieve; however, in practice, real-world multiomics datasets are often incomplete, and multimodal methods that require all modalities to be available restrict learning to a small subset of patients, as illustrated in Figure~\ref{fig:multimodal_motivation}. Although whether multiomics integration consistently improves performance remains an open question, our focus is on scenarios where integration offers improvements over single-modality models. 

\begin{figure}
  \centering
  \includegraphics[width=0.9\linewidth]{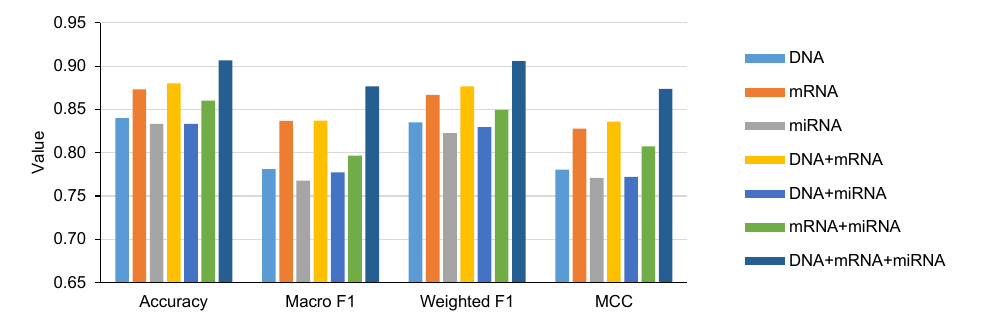}
    \caption{Performance of \texttt{MAGNET}, averaged over five runs, across different combinations of omics modalities on the BRCA dataset.}
  \label{fig:modality_ablation}
\end{figure}

To explore this, we conduct an additional ablation study on the BRCA dataset. Specifically, we select patients who have data available for all three modalities, ensuring that when one or two modalities are removed, the remaining modalities are still present for every patient. We train \texttt{MAGNET} using seven different modality combinations: individual modalities, pairwise combinations of two modalities, and the full set of three modalities. When removing a modality, we mask it for the corresponding patients, treating it as a missing modality. The results of this experiment are shown in Figure~\ref{fig:modality_ablation}. The results show that although mRNA alone has the strongest predictive power among the three modalities, combining it with DNA further improves performance. Moreover, using all three modalities together yields the best overall results across all evaluation metrics.

\begin{figure}
  \centering
  \includegraphics[width=\linewidth]{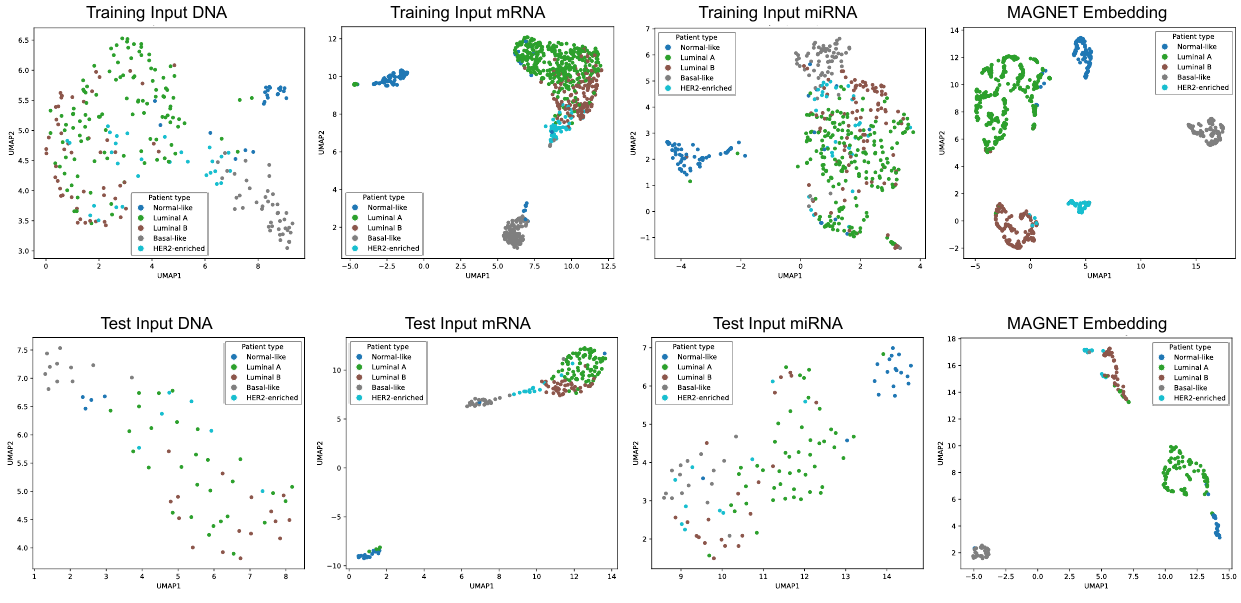}
    \caption{UMAP visualization of the training and test data from the BRCA dataset. For each omics modality, UMAP is generated from the input data, while for \texttt{MAGNET}, it shows the patient representations learned by the GNN module. Top: training data visualization. Bottom: test data visualization.}
  \label{fig:vis}
\end{figure}

To further validate the effectiveness of fusing input omics modalities using \texttt{MAGNET}, we visualize the input omics modalities and the representations learned by the last layer of the GNN module in \texttt{MAGNET} using uniform manifold approximation and projection (UMAP). Figure~\ref{fig:vis} shows the UMAP visualizations for the BRCA dataset, which contains a larger number of patients, providing better clarity compared to the other two datasets. We observe that among the individual omics modalities, mRNA demonstrates greater class separability on both the training and test data, whereas DNA and miRNA provide limited separability, with patients within the same classes overlapping significantly. Specifically, for the mRNA modality, while patients within the Normal-like and Basal-like classes are relatively well separated, there is some overlap between patients in the Luminal A and Luminal B classes. In contrast, \texttt{MAGNET} effectively integrates these modalities, enhancing class separability. The fused representation not only distinguishes patients with similar classes more clearly but also achieves better separation between the Luminal A and Luminal B classes compared to the mRNA modality alone.

\subsection{Separability analysis of patient representations}
\label{app:sep-representation} 
To further evaluate the effectiveness of \texttt{MAGNET}, we measure the separability of patient representations using two clustering metrics: Silhouette Score (SS) \cite{rousseeuw1987silhouettes}, where higher values indicate better-defined clusters, and Davies-Bouldin (DB) index \cite{davies1979cluster}, where lower values reflect better class separation. We extract the representations from the last layer of each trained model before generating predictions. Table~\ref{tbl:vis-base} presents the results on the test data for three datasets.

\begin{table}
  \caption{Separability evaluation of patient representations on test data using Silhouette Score (SS) and Davies-Bouldin (DB) index (\textbf{best}, \underline{second-best}).}
  \label{tbl:vis-base}
  \centering
  \begin{small}
  \begin{tabular}{lcccccc}
        \toprule
            \multicolumn{1}{l}{Method} & \multicolumn{2}{c}{BRCA} & \multicolumn{2}{c}{BLCA} & \multicolumn{2}{c}{OV} \\
            \cmidrule(lr){2-3} \cmidrule(lr){4-5} \cmidrule(lr){6-7}
            & SS (\textbf{\textcolor{teal}{\textuparrow}}) & DB (\textbf{\textcolor{red}{\textdownarrow}}) & SS (\textbf{\textcolor{teal}{\textuparrow}}) & DB (\textbf{\textcolor{red}{\textdownarrow}}) & SS (\textbf{\textcolor{teal}{\textuparrow}}) & DB (\textbf{\textcolor{red}{\textdownarrow}}) \\
        \midrule  
            MOGONET-Zero  & 0.214 & 1.447 & \underline{0.836} & 0.741 & 0.024 & 4.491 \\
            MOGONET-$k$NN & 0.354 & 1.078 & \textbf{0.870} & 0.757 & 0.034 & \underline{4.395} \\
            MRGCN         & 0.145 & 1.658 & 0.261 & 0.700 & 0.006 & 8.808 \\
            M3Care        & \underline{0.437} & \underline{0.826} & 0.330 & \underline{0.670} & 0.040 & 4.551 \\
            MUSE          & 0.408 & 0.879 & 0.295 & 0.793 & \underline{0.047} & \textbf{3.900} \\
        \midrule  
            \texttt{MAGNET} & \textbf{0.440} & \textbf{0.789} & 0.578 & \textbf{0.642} & \textbf{0.048} & 4.433 \\
        \bottomrule
  \end{tabular}
  \end{small}
\end{table}

On BRCA, \texttt{MAGNET} improves SS and reduces DB, demonstrating its ability to separate classes effectively in a multiclass classification task. On BLCA, an imbalanced dataset, \texttt{MAGNET} achieves the lowest DB index, indicating strong global class separation. However, its SS is moderate, likely due to minority class patients being closer to decision boundaries. In contrast, imputation-based methods yield a higher SS by estimating missing values based on dominant patterns, which results in more compact clusters and improved local cohesion. On OV, a well-balanced dataset, \texttt{MAGNET} achieves the highest SS, highlighting well-clustered patient representations. However, its DB index remains average, likely because balanced class distributions result in uniform cluster compactness, reducing the inter-cluster contrast that DB emphasizes.

These results show \texttt{MAGNET}'s ability to enhance class separability across datasets while balancing local cohesion and global structure in learned representations.

We next evaluate eight types of representations used in the \texttt{MAGNET} architecture by measuring the separability of patient representations in the test data using SS and DB index: three input modalities, their learned representations from omics-specific encoders, the fused representation from PMMHA, and the final-layer representation produced by the GNN. Table~\ref{tbl:vis-abl} presents the results, showing that in all cases across the three datasets, the final-layer representation learned by \texttt{MAGNET} achieves the highest separability. Moreover, the representation learned from PMMHA alone achieves nearly the second-best result across all datasets compared to individual omics representations, suggesting that each added component in \texttt{MAGNET} contributes to producing more informative representations. 

\begin{table}
  \caption{Separability evaluation of patient representations learned from \texttt{MAGNET} modules on test data using Silhouette Score (SS) and Davies-Bouldin (DB) index (\textbf{best}, \underline{second-best}).}
  \label{tbl:vis-abl}
  \centering
  \begin{small}
  \begin{tabular}{lcccccc}
    \toprule
        \multicolumn{1}{l}{Module} & \multicolumn{2}{c}{BRCA} & \multicolumn{2}{c}{BLCA} & \multicolumn{2}{c}{OV} \\
        \cmidrule(lr){2-3} \cmidrule(lr){4-5} \cmidrule(lr){6-7}
        & SS (\textbf{\textcolor{teal}{\textuparrow}}) & DB (\textbf{\textcolor{red}{\textdownarrow}}) & SS (\textbf{\textcolor{teal}{\textuparrow}}) & DB (\textbf{\textcolor{red}{\textdownarrow}}) & SS (\textbf{\textcolor{teal}{\textuparrow}}) & DB (\textbf{ \textcolor{red}{\textdownarrow}}) \\
    \midrule  
        Input DNA       & 0.035 & 2.635 & 0.231 & 0.987 & 0.002 & 7.833 \\
        Input mRNA      & 0.105 & 2.276 & 0.120 & 1.531 & 0.006 & 5.264 \\
        Input miRNA     & 0.046 & 3.307 & 0.024 & 2.331 & 0.005 & 6.961 \\
        \midrule
        DNA Embedding   & 0.017 & 2.522 & 0.264 & 0.977 & 0.002 & 7.807 \\
        mRNA Embedding  & 0.246 & 1.170 & 0.188 & 1.284 & \underline{0.018} & \underline{4.618} \\
        miRNA Embedding & 0.018 & 3.571 & 0.004 & 2.421 & 0.006 & 6.897 \\
        \midrule
        PMMHA Embedding & \underline{0.295} & \underline{0.983} & \underline{0.345} &  \underline{0.865} & 0.009 & 6.585 \\
        GNN Embedding   & \textbf{0.440} & \textbf{0.789} & \textbf{0.578} & \textbf{0.642} & \textbf{0.048} & \textbf{4.433} \\
    \bottomrule
  \end{tabular}
  \end{small}
\end{table}

\subsection{Sensitivity analysis}
\label{app:sensitivity_analysis} 

\begin{figure}
  \centering
  \includegraphics[width=0.7\linewidth]{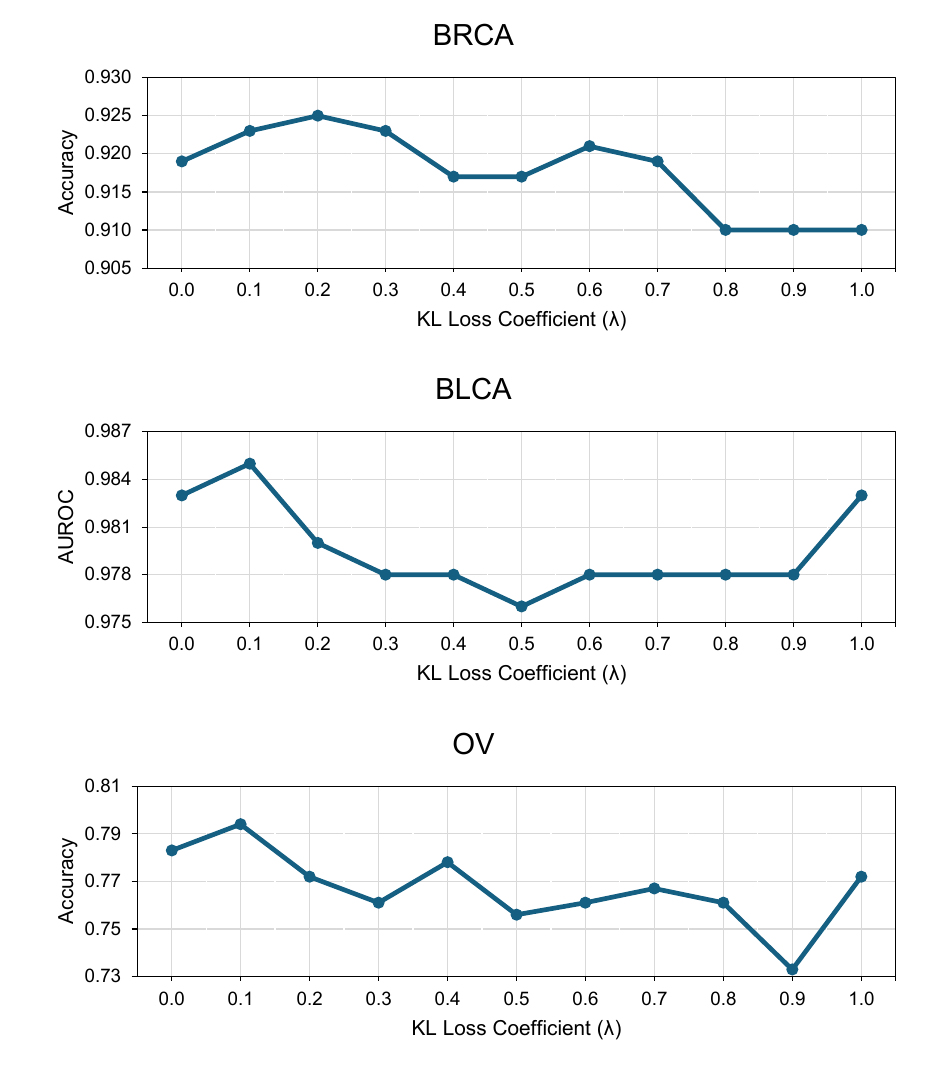}
    \caption{Impact of the KL loss coefficient ($\lambda$) on \texttt{MAGNET} performance over validation sets.}
  \label{fig:sensitivity_loss}
\end{figure}

We analyze the sensitivity of \texttt{MAGNET} to two key hyperparameters: the KL loss coefficient ($\lambda$) and the graph sparsity rate. First, we evaluate the impact of the KL loss coefficient by varying $\lambda \in [0,1]$, where $\lambda = 0$ represents training without the KL loss (Figure~\ref{fig:sensitivity_loss}). The results show that performance varies smoothly across different values of $\lambda$, with the best performance observed around $\lambda = 0.1$. Compared to $\lambda = 0$, incorporating the KL loss improves performance, which is consistent with the ablation results reported in Table~\ref{tbl:abl-cls}. As $\lambda$ increases further, performance gradually decreases, indicating that excessively large values may reduce the contribution of the supervised objective.

\begin{figure}
  \centering
  \includegraphics[width=\linewidth]{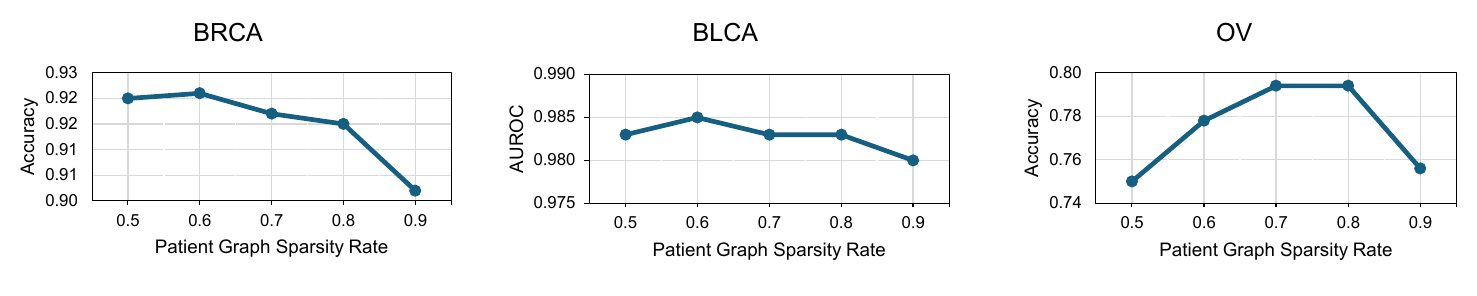}
    \caption{Impact of patient graph sparsity rate on \texttt{MAGNET} performance over validation sets.}
  \label{fig:sensitivity_sparse}
\end{figure}

We also analyze the impact of the graph sparsity rate by varying it within the range $[0.5, 0.9]$ (Figure~\ref{fig:sensitivity_sparse}). The results show that performance remains relatively stable across different sparsity rates, with variation below 2\% for BRCA and BLCA. The OV dataset shows slightly higher sensitivity ($\sim$4.5\%) but follows a similar trend. We observe that high sparsity levels can reduce performance, suggesting that the graph becomes too disconnected for effective information sharing, while very low sparsity levels may introduce noise from less similar patients.

\subsection{Learned attention weight analysis}
\label{app:att_weight} 

\begin{figure}
  \centering
  \includegraphics[width=\linewidth]{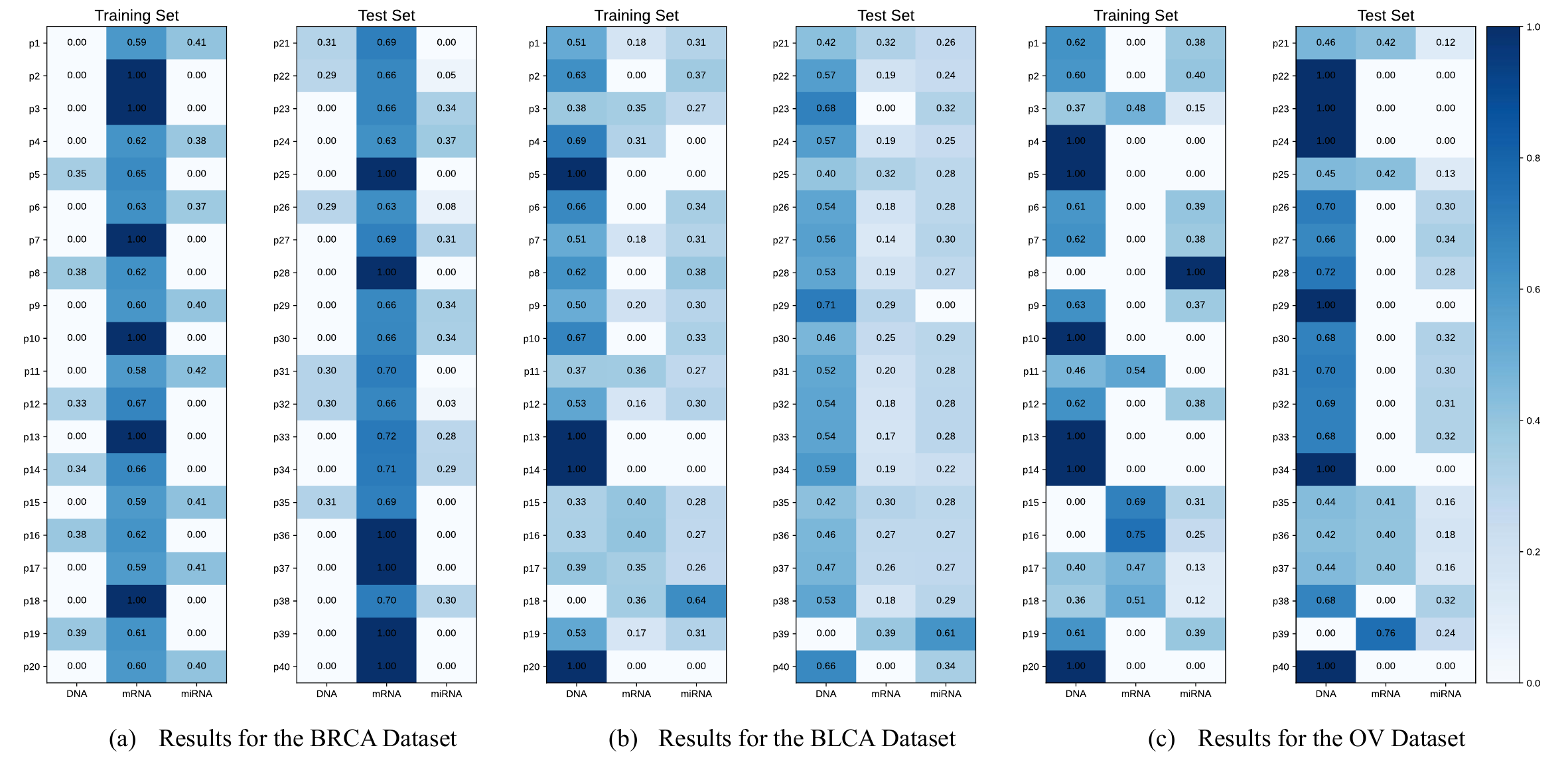}
    \caption{Attention weights learned by \texttt{MAGNET} for 20 selected patients across modalities in training and test sets.}
  \label{fig:att_weight}
\end{figure}

In order to verify whether modality importance varies across patients or subtypes, we conduct three analyses of the learned attention weights. First, we visualize attention weights for 20 representative patients across all datasets (Figure~\ref{fig:att_weight}). The heatmaps show clear variability in attention across patients and modalities, indicating that the model assigns different importance depending on each patient’s profile and the available information. These results demonstrate that PMMHA learns patient-specific modality importance rather than fixed global weights.

\begin{figure}
  \centering
  \includegraphics[width=\linewidth]{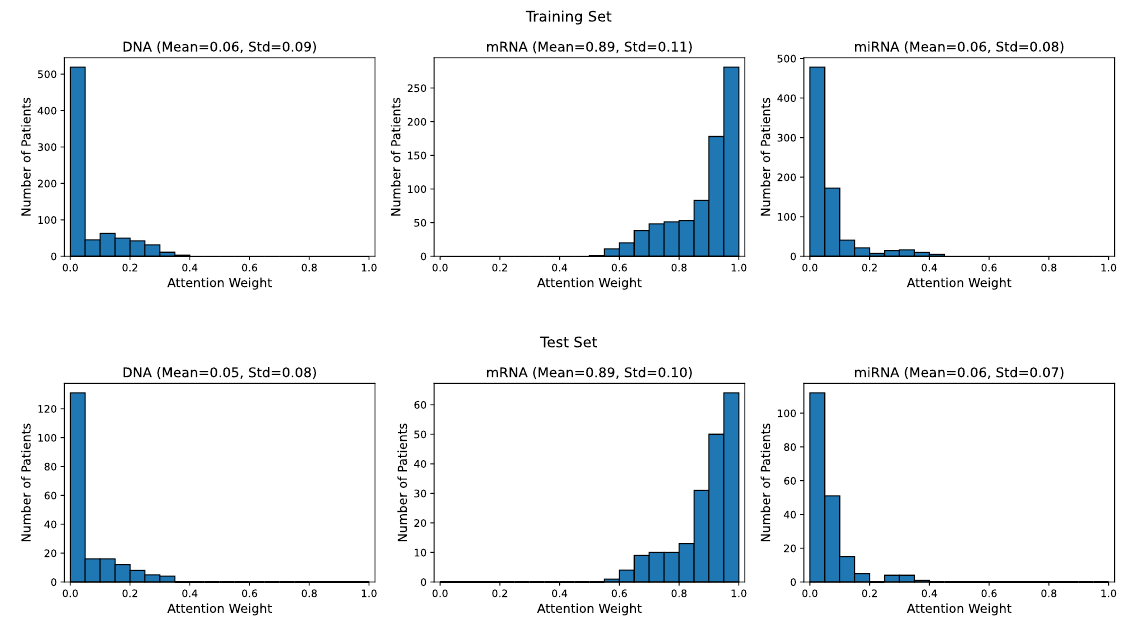}
    \caption{Distribution of attention weights learned by \texttt{MAGNET} across modalities for the BRCA dataset in training and test sets. Mean and standard deviation are reported for each modality.}
  \label{fig:att_modality_brca}
\end{figure}

\begin{figure}
  \centering
  \includegraphics[width=\linewidth]{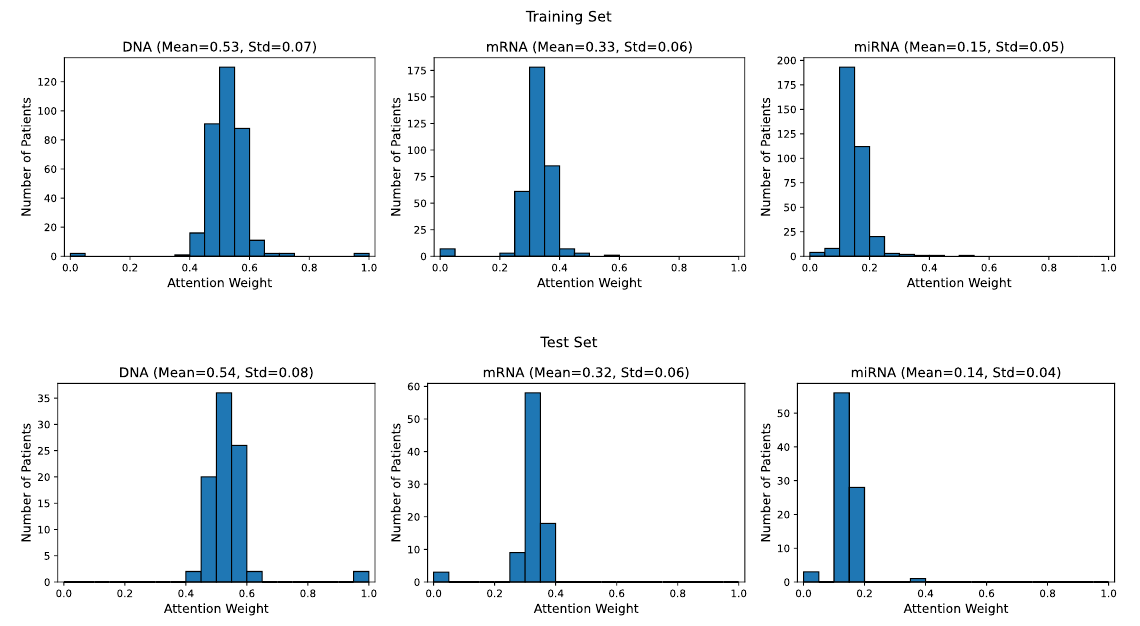}
    \caption{Distribution of attention weights learned by \texttt{MAGNET} across modalities for the BLCA dataset in training and test sets. Mean and standard deviation are reported for each modality.}
  \label{fig:att_modality_blca}
\end{figure}

\begin{figure}
  \centering
  \includegraphics[width=\linewidth]{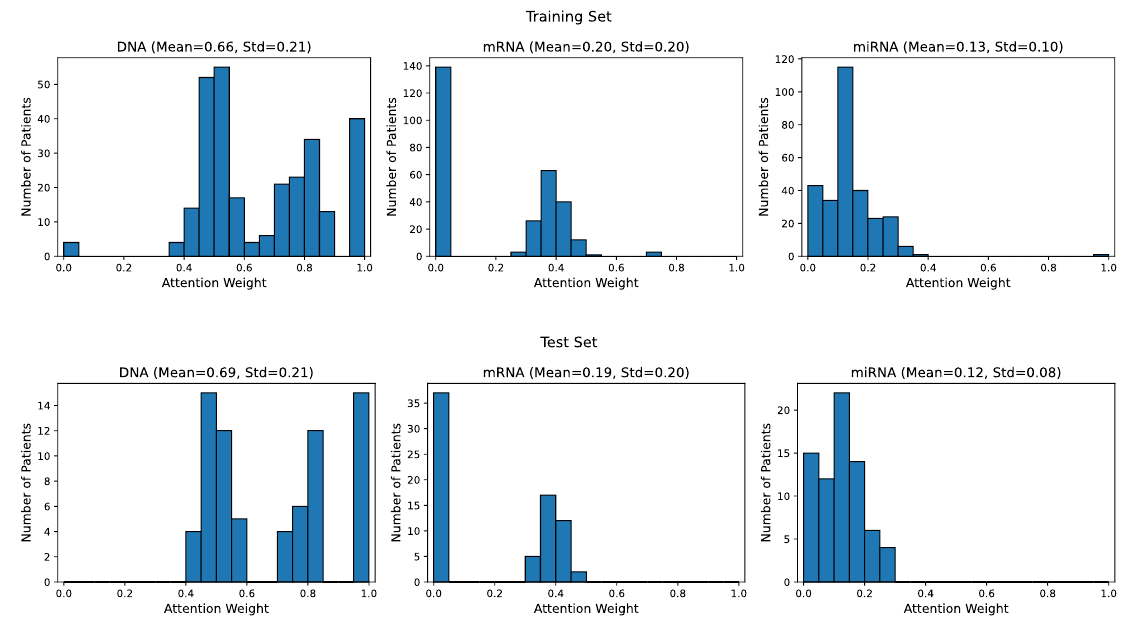}
    \caption{Distribution of attention weights learned by \texttt{MAGNET} across modalities for the OV dataset in training and test sets. Mean and standard deviation are reported for each modality.}
  \label{fig:att_modality_ov}
\end{figure}

We then analyze the distribution of attention weights across all patients for each modality (Figures~\ref{fig:att_modality_brca}--\ref{fig:att_modality_ov}). For BRCA, mRNA consistently receives the highest attention, which aligns with its complete availability across patients. For BLCA, DNA receives relatively higher attention, while mRNA and miRNA still contribute meaningfully, reflecting a more balanced integration due to lower missingness. For OV, DNA shows the strongest contribution, followed by miRNA, consistent with their lower missingness rates. These results highlight PMMHA’s ability to adjust weights under heterogeneous missing-modality patterns.

\begin{figure}
  \centering
  \includegraphics[width=\linewidth]{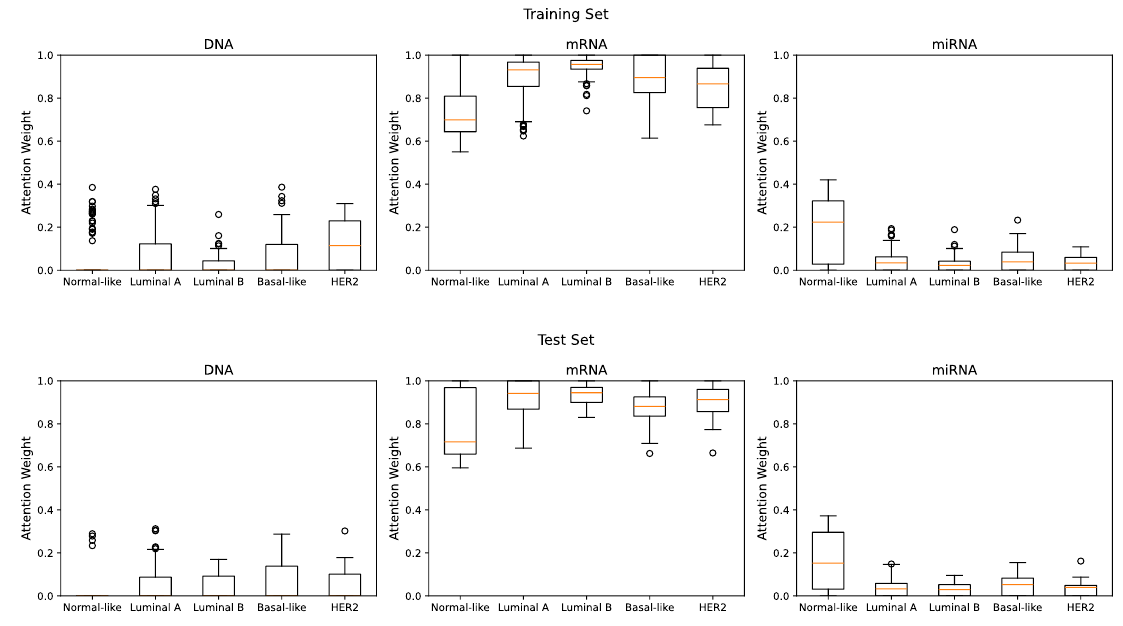}
    \caption{Attention weight distributions learned by \texttt{MAGNET} across modalities for different classes in the BRCA dataset, shown for training and test sets.}
  \label{fig:att_class_brca}
\end{figure}

\begin{figure}
  \centering
  \includegraphics[width=\linewidth]{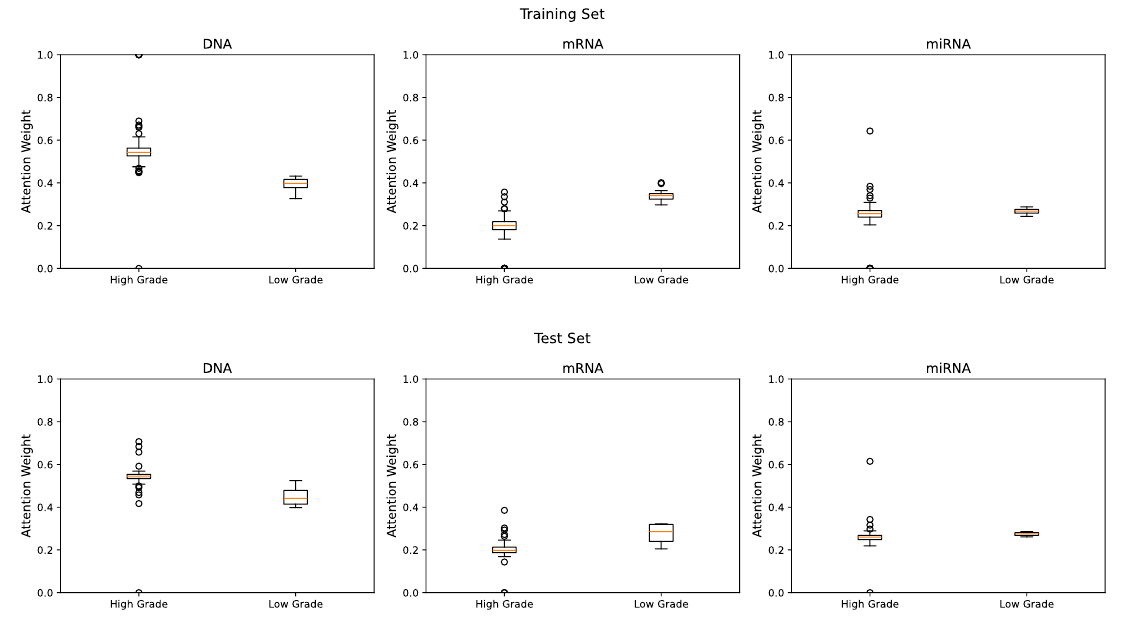}
    \caption{Attention weight distributions learned by \texttt{MAGNET} across modalities for different classes in the BLCA dataset, shown for training and test sets.}
  \label{fig:att_class_blca}
\end{figure}

\begin{figure}
  \centering
  \includegraphics[width=\linewidth]{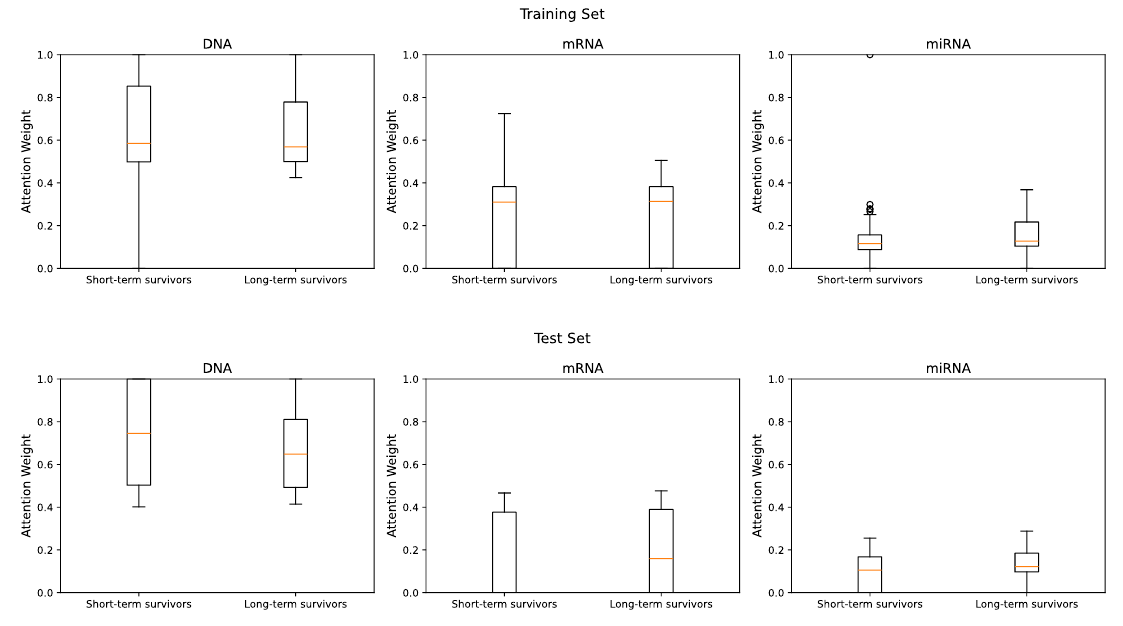}
    \caption{Attention weight distributions learned by \texttt{MAGNET} across modalities for different classes in the OV dataset, shown for training and test sets.}
  \label{fig:att_class_ov}
\end{figure}

Finally, we analyze attention weight distributions across different classes within each dataset (Figures~\ref{fig:att_class_brca}--\ref{fig:att_class_ov}). The results show variation in attention patterns across classes. For example, in the BRCA dataset, different subtypes exhibit distinct distributions of modality weights. Similarly, in the BLCA and OV datasets, we observe shifts in attention distributions between classes. These results provide additional evidence that PMMHA does not rely on fixed modality weights and can flexibly adjust modality importance across patients.

\subsection{Patient graph connectivity analysis}
\label{app:graph_analysis} 

In multiomics settings, different modalities (e.g., DNA vs. mRNA) capture distinct biological aspects and are not directly comparable. \texttt{MAGNET} therefore connects patients only when they share at least one modality, ensuring comparisons within a consistent context. This results in a biologically meaningful graph construction strategy that explicitly accounts for missing modalities. Moreover, isolated nodes are connected to their most similar neighbor to preserve graph connectivity. To justify these design choices, we analyze the constructed patient graph from three perspectives.

\begin{figure}
  \centering
  \includegraphics[width=\linewidth]{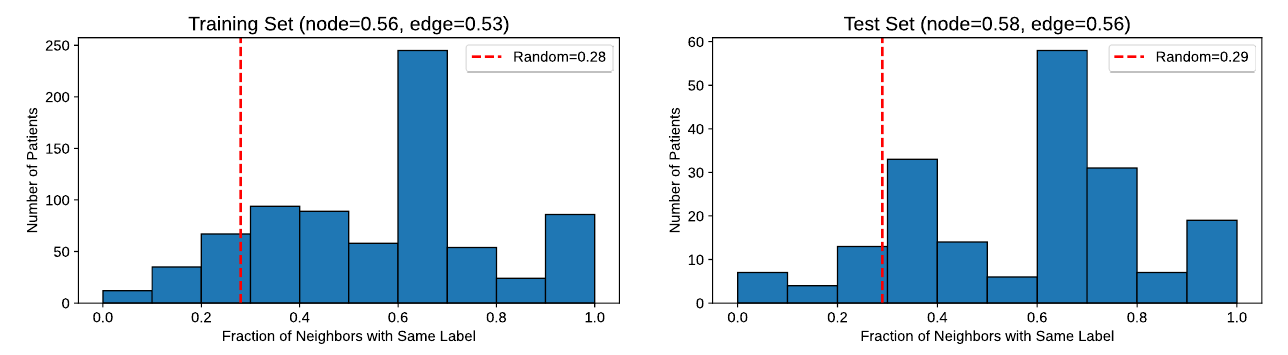}
    \caption{Label homophily analysis of the \texttt{MAGNET} patient graph on the BRCA dataset with five subtypes. The red dashed line indicates the random baseline based on class proportions.}
  \label{fig:node_edge_homophily}
\end{figure}

\begin{figure}
  \centering
  \includegraphics[width=\linewidth]{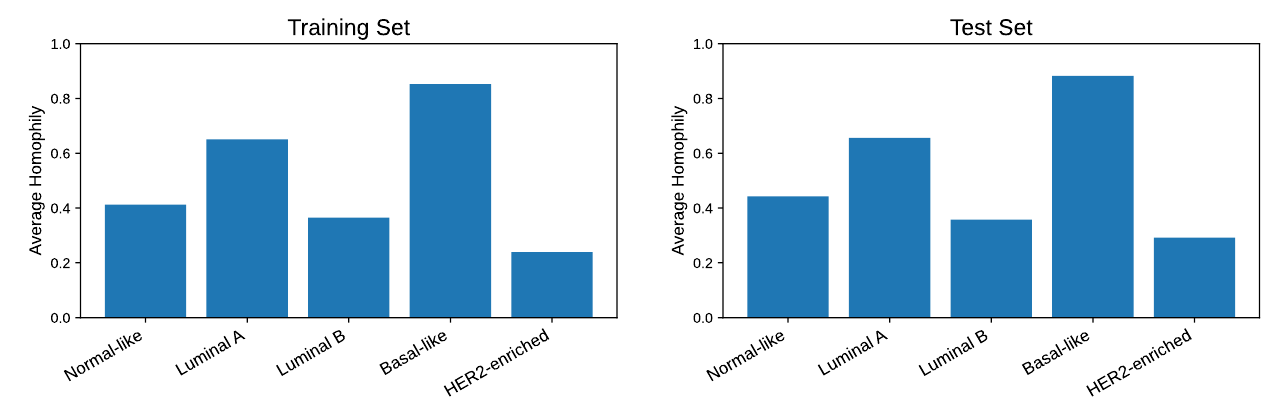}
    \caption{Label homophily of \texttt{MAGNET} across subtypes on the BRCA dataset with five subtypes.}
  \label{fig:homophily_class}
\end{figure}

First, to provide insight into the graph structure, we analyze node and edge homophily, two standard graph properties, on the BRCA dataset, which contains the largest number of patients (Figures~\ref{fig:node_edge_homophily}--\ref{fig:homophily_class}). Figure~\ref{fig:node_edge_homophily} shows that both node and edge homophily are approximately twice as high as a random baseline (0.56 vs.~0.28 and 0.53 vs.~0.28 for training, and 0.58 vs.~0.29 and 0.56 vs.~0.29 for test), indicating that the graph captures label-consistent relationships. Figure~\ref{fig:homophily_class} further shows consistent homophily across subtypes, suggesting that the graph structure is not dominated by a single class. These results provide direct evidence that the constructed patient graph forms meaningful neighborhoods for message passing, supporting predictions based on clinically relevant patient similarities.

\begin{figure}
  \centering
  \includegraphics[width=\linewidth]{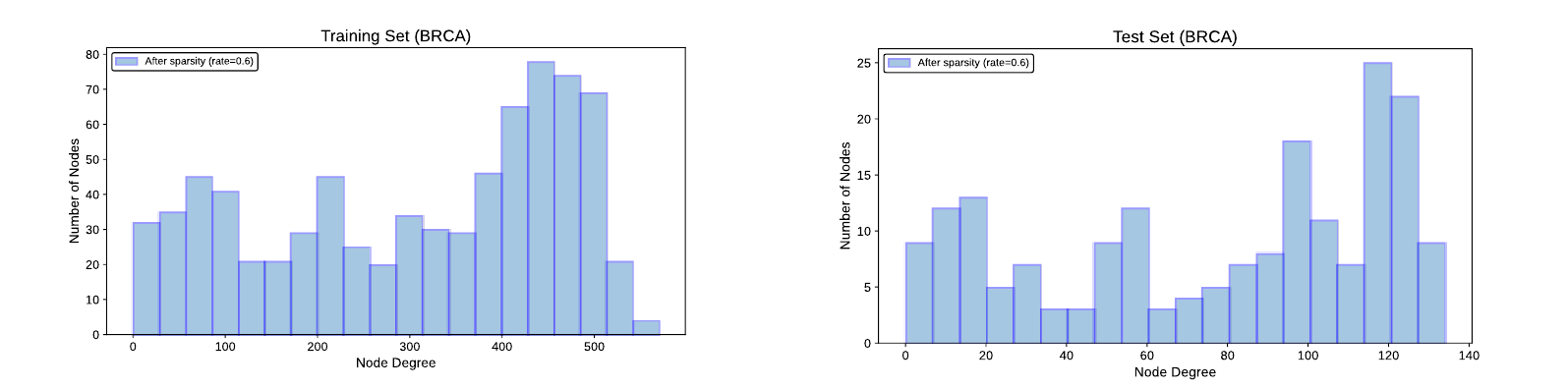}
    \caption{Node degree distributions after applying graph sparsity in \texttt{MAGNET} on the BRCA dataset across training and test sets. Isolated nodes are not reconnected after sparsity. Sparsity rates are selected via hyperparameter tuning (Table~\ref{tbl:hyp}).}
  \label{fig:node_degree_blca}
\end{figure}

Second, to evaluate the robustness of the patient graph to potentially noisy edges, we analyze node degree distributions after graph sparsification (Figure~\ref{fig:node_degree_blca}). The results show that each node retains a sufficient number of neighbors after filtering, reducing the influence of noisy edges through neighborhood aggregation in the GNN.

\begin{figure}
  \centering
  \includegraphics[width=0.9\linewidth]{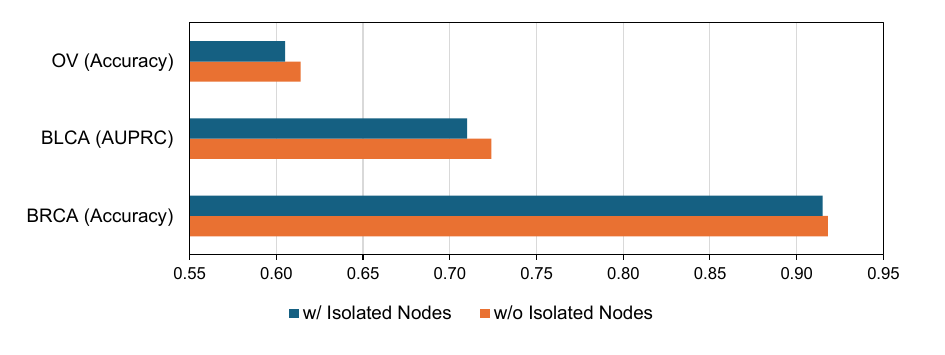}
    \caption{Ablation study on the effect of handling isolated nodes by connecting them to the most similar patient in \texttt{MAGNET}.}
  \label{fig:abl_isolation}
\end{figure}

Finally, we conduct an ablation study comparing performance with and without reconnecting isolated nodes (Figure~\ref{fig:abl_isolation}). The results show that reconnecting isolated nodes consistently improves performance across all datasets. This is because isolated nodes cannot participate in message passing, whereas reconnecting them to their most similar patients enables them to receive informative signals. Since these connections are formed using the most similar patients rather than arbitrary ones, they are more likely to preserve meaningful patient relationships.

\subsection{Execution time analysis}
\label{app:time-analysis} 
\begin{figure}
  \centering
  \includegraphics[width=\linewidth]{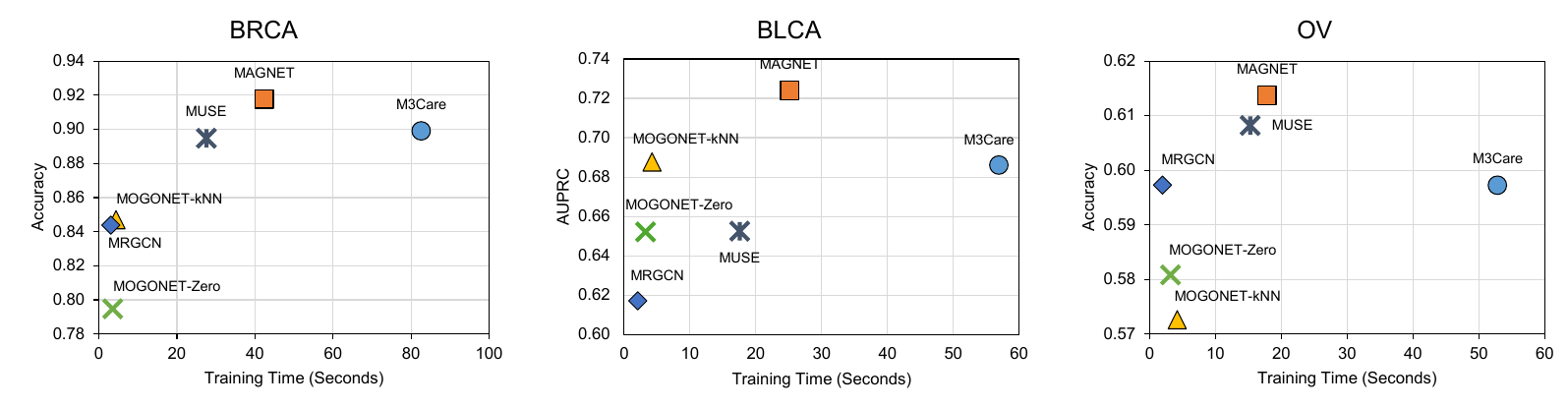}
    \caption{Comparison of classification performance and execution time, averaged over five runs.}
  \label{fig:time_complexity}
\end{figure}

We first evaluate the execution time versus classification performance of all methods during training on the BRCA, BLCA, and OV datasets. The results, averaged over five runs, are presented in Figure~\ref{fig:time_complexity}. \texttt{MAGNET} achieves the best prediction performance among all methods, although its execution time is relatively higher than some baselines. This demonstrates that \texttt{MAGNET} offers a favorable trade-off between predictive performance and computational efficiency. Among the baselines, \texttt{MAGNET} execution time is slightly higher than that of the recent method MUSE. In contrast, M3Care as an imputation-based method shows the worst computational time which is almost double that of \texttt{MAGNET}. This highlights the high computational cost of imputation-based approaches. The lightweight computational time of MOGONET-Zero and MOGONET-$k$NN is attributed to the fact that their imputation is a simple procedure performed before training. However, this simplicity comes at the cost of significantly worse classification performance in almost all cases.

\begin{figure}
  \centering
  \includegraphics[width=\linewidth]{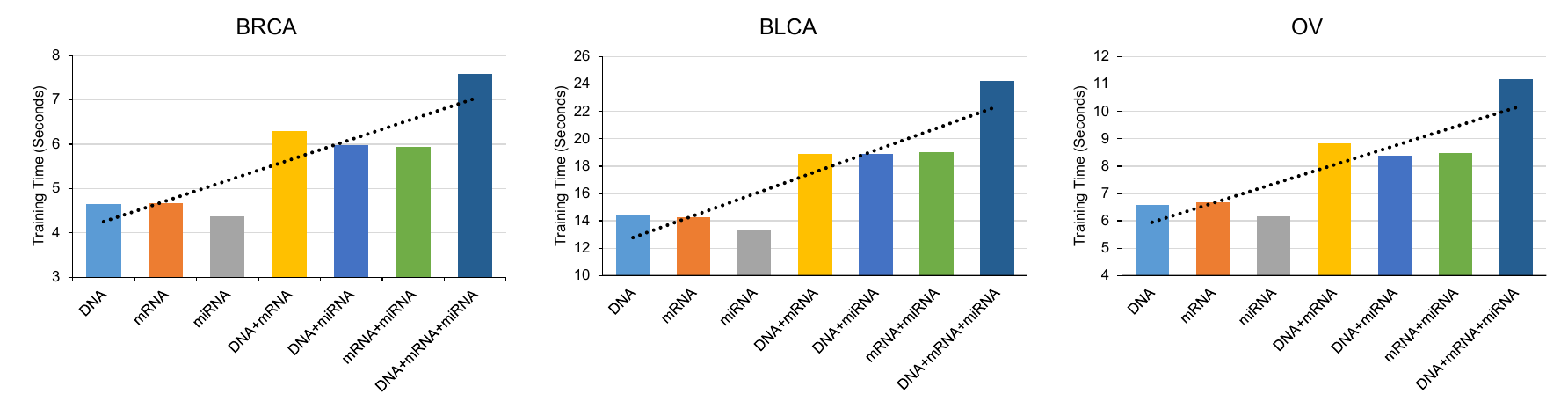}
    \caption{Average execution time of \texttt{MAGNET} over five runs across different omics modality combinations.}
  \label{fig:time_ablation}
\end{figure}

We next present the execution time of \texttt{MAGNET} with an increasing number of input modalities. To conduct this experiment, we retain only patients with all modalities available, ensuring that all modality combinations have the same number of patients, which allows a fair comparison. We report the average execution time over five runs for different combinations of modalities, and the results are shown in Figure~\ref{fig:time_ablation}. We observe that, across all datasets, the runtime increases approximately linearly as more modalities are added. This indicates that the computational cost of \texttt{MAGNET} scales linearly with the number of input modalities.

\end{document}